\pdfoutput=1
\documentclass[shortAfour, sageh, times]{sagej}
\input{preprint_formatting.tex}
\usepackage[utf8]{inputenc}

\makeatletter 
\renewcommand\subsubsection{\@startsection{subsubsection}{3}{\z@}%
                                     {0.9\@bls plus .3\@bls minus .1\@bls}%
                                     {4pt\@afterindentfalse}%
                                     {\sagesf\large\itshape\raggedright}}
\makeatother

\usepackage{xcolor} 
\usepackage{hyperref} 
\usepackage{tabularx} 
\usepackage{float} 

\usepackage[subrefformat=parens,labelformat=parens]{subfig} 

\usepackage{etoolbox}
\apptocmd{\sloppy}{\hbadness 10000\relax}{}{}

    \newcommand{\new}[1]{#1}
    \newcommand{\newoutline}[1]{#1}
    \newcommand{\cut}[1]{}
    \newcommand{\cutcitation}[1]{}

\DeclareMathOperator*{\argmin}{arg\,min}

\def\calO{{\mathcal O}}
\def\calE{{\mathcal E}}
\def\Rot{{\cal R}}
\def\rot{{\tt R}}
\def\norm#1{\left\|{#1}\right\|}

\def\tr{^T}
\def\sq{^2}

\def\rr#1{\mathbb{R}^{#1}}

\begin{document}
\runninghead{Elias and Wen}

\title{Redundancy parameterization and inverse kinematics of 7-DOF revolute manipulators}
\author{Alexander~J.~Elias and John~T.~Wen}
\affiliation{Department of Electrical, Computer, and Systems Engineering, Rensselaer Polytechnic Institute, Troy, NY}
\corrauth{Alexander J. Elias, Rensselaer Polytechnic Institute, Troy, NY.}

\email{eliasa3@rpi.edu}

\begin{abstract} 
Seven degree-of-freedom (DOF) robot arms have one redundant DOF which does not change the\cut{ translational or rotational} motion of the end effector.
The redundant DOF offers greater manipulability of the arm configuration to avoid obstacles and \cut{steer away from} singularities, but it must be parameterized to fully specify the joint angles for a given end effector pose.
For 7-DOF revolute (7R) manipulators, we introduce a new concept of generalized shoulder-elbow-wrist (SEW) angle, a generalization of the conventional SEW angle but with an arbitrary choice of the reference direction function.
The SEW angle is \new{widely used and} easy for human operators to visualize as a rotation of the elbow about the \new{shoulder-wrist line}\cut{line from the shoulder to the wrist and has been used in the teleoperation of space robot arms}.
\cut{Since the conventional SEW angle formulation is prone to singularities, we introduce a special choice of the reference direction function called the stereographic SEW angle which has a singularity in only one direction in the workspace.}%
\new{Since other redundancy parameterizations including the conventional SEW angle encounter an algorithmic singularity along a line in the workspace, we introduce a special choice of the reference direction function called the stereographic SEW angle which has a singularity only along a half-line, which can be placed out of reach.}
We prove that such a singularity is unavoidable for any parameterization.
We also include expressions for the SEW angle Jacobian along with singularity analysis.
Finally, we provide \new{efficient and singularity-robust} inverse kinematics solutions for most known 7R manipulators using the general SEW angle and the subproblem decomposition method.
These solutions are often closed-form but may sometimes involve a 1D or 2D search \new{in the general case}.
\new{Search-based solutions may be converted to finding zeros of a high-order polynomial.}
Inverse kinematics solutions, examples, and evaluations are available in a publicly accessible repository.
\end{abstract}


\maketitle

\section{Introduction}
\begin{figure}[t]
    \centering
    \includegraphics[scale = 0.5, clip]{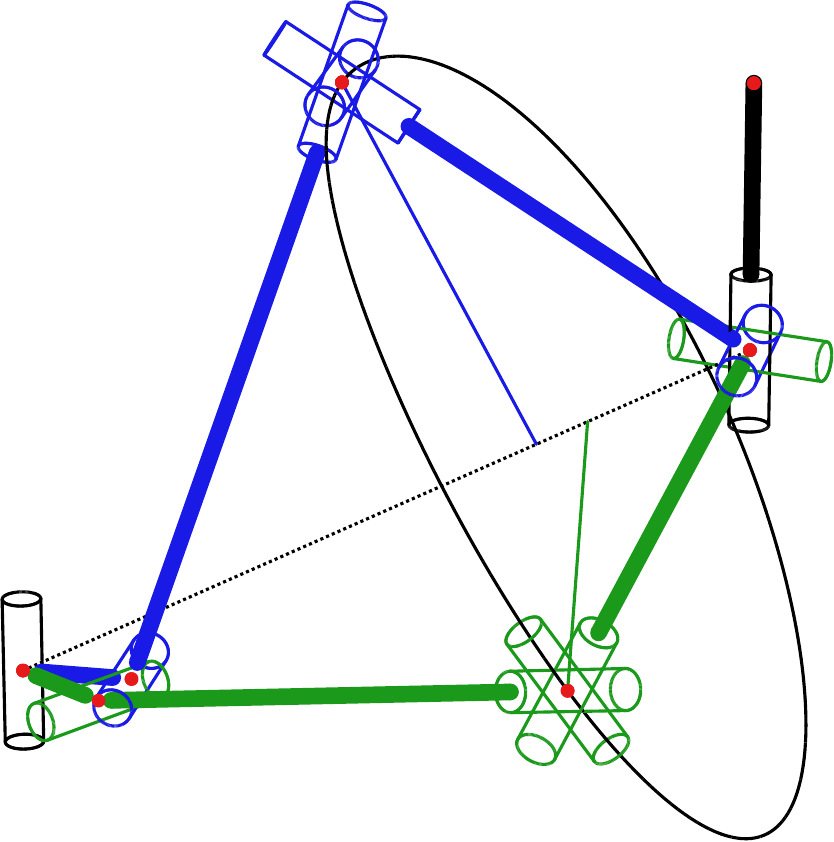}
    \caption{Self-motion for a Motoman SIA50D (R-R-3R\textsuperscript{E}-2R) robot arm. For a given end effector pose, the spherical elbow may be on a curve which is the intersection of a torus formed by joints 1 and 2 a sphere centered at joints 6 and 7.}
    \label{fig:manifold}
\end{figure}

Most industrial robot arms have six revolute joints to control the six degrees of freedom (DOF) of the robot end effector pose,
but a human arm has seven DOF: three for the  shoulder, one for the elbow, and three for the wrist. Similarly, there are 7-DOF revolute (7R) robot arms such as the
    Robotics Research Corporation (RRC) arm \citep*{RRC},
    Baxter \citep*{baxter},
    Sawyer \citep*{sawyer},
    Yumi \citep*{ABB_YUMI},
    and the Space Station Robot Manipulator System (SSRMS) \citep*{crane1991kinematic}.
The extra {\em redundant} degree of freedom means that there is a continuum of arm configurations for a given hand or robot end effector pose. Holding the end effector pose constant while moving through the continuum of arm configurations, called self-motion, commonly looks like the elbow rotating around the line passing from the shoulder to the wrist (Figure~\ref{fig:manifold}).
Benefits of 7R arms over 6R arms include using redundancy to
    avoid singularities and obstacles \citep*{hollerbach1985optimum},
    optimize motion time \citep*{chen2023kinematics},
    avoid joint motion limits \citep*{flacco2012motion},
    and avoid joint torque limits \citep*{hollerbach1987redundancy}.

To fully specify the pose of a 7R robot up to a finite number of solutions, the end effector pose must be augmented by a secondary task.
During pure self-motion, the only difference between parameterizations would be the rate of movement; differences between parameterizations are made clearer during motion of the end effector.
Redundancy in human and many 7R robot arms may be conveniently parameterized by the shoulder-elbow-wrist (SEW) angle, sometimes called the elbow angle, which characterizes the rotation of the plane containing the shoulder, elbow, and wrist about the shoulder-wrist line with respect to a reference plane. This reference plane is conventionally chosen as the plane containing the shoulder-wrist line and a reference vector. This redundancy parameterization is easy to visualize
\new{and is widely used in applications such as space teleoperation
\citep*{tsumaki2001numerical, pryor2023teleoperation, zhu2021design}
partly because it is intuitive for space robot operators \citep*{swaim1994use}}.
Furthermore, for certain 7R arms, the inverse kinematics has an analytical solution, i.e., for a given robot end effector pose and SEW angle, the finite set of the seven robot joint angles may be solved directly instead of iteratively. The augmented Jacobian, the \(7\times7\) matrix that maps the joint velocity vector to the end effector spatial velocity and the SEW angular velocity, is easily characterized.

An issue with the conventional SEW angle becomes apparent when the shoulder-wrist line is collinear with the reference vector because the reference plane becomes undefined. This is referred to as an {\em algorithmic singularity}, a singularity due to the choice of redundancy parameterization.
\new{
Algorithmic singularities are the main weakness of parameterizations like the SEW angle \citep*{dupuis2001general} because robots experience undesirable behavior near them just as for kinematic singularities. For example, being near an algorithmic singularity may result in dangerously large elbow movement, which is is especially problematic for teleoperation \citep*{tsumaki2001numerical}. It also leads to poor convergence for iterative algorithms as well as numerical precision problems since many significant digits are required to accurately specify the robot pose.
Designers must carefully choose the reference direction to avoid singularities, but it is often inevitable to waste regions of the reachable workspace \citep*{carignan2000partitioned, carignan2001controlling, naylor2007visual, scott2008line}.
}

In this paper, we introduce a new generalized concept of SEW angle which includes the conventional SEW angle as a special case.  We show that an algorithmic singularity is unavoidable for any redundancy parameterization, but a special choice of the generalized SEW angle based on the stereographic projection changes the bidirectional singularity to a unidirectional singularity. The singularity direction may be located towards the base of the arm so it will not be encountered in the robot workspace. The advantages of the SEW angle redundancy parameterization are still retained for the generalized SEW angle, including analytical inverse kinematics for many 7R arms, intuitive teleoperation, and an analytical Jacobian. \new{Most existing algorithms which use the conventional SEW angle or its Jacobian can be easily adapted to use the general SEW angle, including the stereographic SEW angle}

We also provide inverse kinematics (IK) solutions using the general SEW angle \cut{and the subproblem decomposition method \mbox{\citep*{elias2024canonical}}} for most 7R robots used in practice or mentioned in the literature \new{including general 7R manipulators, as well as arms which do not yet exist.
These solutions are built upon IK-Geo, an open-source and easy-to-use IK solver based on a unifying subproblem decomposition approach \citep*{elias2024canonical}.}
The solutions are often closed-form, but for some robots IK is solved using a 1D or 2D search\new{{ over a compact set}}.
\new{
These analytical and semi-analytical methods are efficient, precise, and stable, especially compared to Jacobian-based algorithms. 
They are also robust to singularities and find all IK solutions, including singular solution, rather than just one close to an initial guess. For branches without exact solutions, they return continuous non-exact and sometimes least-squares solutions. This allows Cartesian motion directly through singularities to switch between different IK branches without resorting to joint control.
We also demonstrate converting a search-based solution to a polynomial solution: To the best of our knowledge, we are the first to provide a polynomial in the tangent half-angle of one joint which corresponds to IK solutions of a 7R robot parameterized by SEW angle.}

\new{
We offer the following new contributions in this paper:
\begin{itemize}
\item We introduce the stereographic SEW angle, a new redundancy parameterization that encounters a singularity only along a half-line instead of a full line as in the conventional SEW angle. This significantly enlarges of the singularity-free region of the workspace.
\item We show that algorithmic singularities are unavoidable for any redundancy parameterization. We classify different types of singularities for 7R arms and discuss when singularities are concurrent.
\item We efficiently solve IK for any 7R manipulator parameterized by the conventional or stereographic SEW angle using the subproblem decomposition approach. We categorize 7R robots based on intersecting or parallel axes and choices of shoulder, elbow, and wrist locations. We also demonstrate solving IK for a 7R manipulator parameterized by conventional or stereographic SEW angle by finding a high-order polynomial in the tangent half-angle of one joint.
\end{itemize}
}

The remainder of the paper is organized as follows.
In Section~\ref{sec:background} we discuss previous related works, and we describe forward kinematics for 7R arms using coordinate-free notation and the product of exponentials approach.
We introduce the general SEW angle in Section~\ref{sec:gen_SEW} including forward and differential kinematics, we  discuss singularity conditions, and we relate it to the conventional SEW angle.
In Section~\ref{sec:singularityexistence} we prove the existence of algorithmic singularities for any redundancy parameterization.
We define the stereographic SEW angle in Section~\ref{sec:stereographic_SEW_angle}, discuss its relationship to stereographic projection and its singularity behavior, and provide an example comparing using the conventional and stereographic SEW angle formulations.
We provide IK solutions \new{with examples }in Section~\ref{sec:IK}, and we conclude in Section~\ref{sec:conclusion}.

Inverse kinematics solutions, examples, and evaluations are available in a publicly accessible repository\footnote{\url{https://github.com/rpiRobotics/stereo-sew}}.

\section{Background}\label{sec:background}
\subsection{Robot Notation}
To notate different families of robot kinematic parameters, we follow and slightly extend the notation introduced by \cite*{pieper1969kinematics}. A single revolute joint is notated by R, and when multiple joints intersect, they may be notated as 2R or 3R for two or three intersecting joints, respectively. The joints are written in order from the base to the end effector. For example, a robot with a spherical shoulder, a revolute elbow, and a spherical wrist may be notated as \mbox{3R-R-3R}. We also introduce the notation of 2R\textbar\textbar{} and 3R\textbar\textbar{} to indicate two or three consecutive parallel revolute joints. Note that since a given joint may intersect or be parallel with both the joint before it and the joint after it, a single robot may fall into multiple robot kinematic families.

We use superscript S, E, or W to indicate when the shoulder, elbow, or wrist is placed at a joint or joint intersection. For example, 2R\textsuperscript{E} would indicate a 2R joint with the elbow placed at the joint intersection. Unless otherwise indicated, a robot has its shoulder on the first joint or joint intersection and has its wrist on its last joint or joint intersection.

\subsection{Related Literature}
The analysis and parameterization of the redundant degree of freedom in 7R manipulators has been primarily driven by space robotics applications and more recently motivated by industrial robots in manufacturing and humanoid robots.
Early papers focused on analyzing the geometry, topology, and differential kinematics of 7-DOF manipulators. \cite*{baillieul1985kinematic} introduced the concept of the augmented Jacobian (termed the extended Jacobian) and discussed the algorithmic singularity. \cite*{hollerbach1985optimum} discussed options for 7-DOF manipulator designs, including how adding an extra degree of freedom removes internal singularities. The paper discusses the self-motion manifold and was one of the earliest papers to recommend using the conventional SEW angle (with the reference vector pointing up). \cite*{burdick1989characterization} gave an overview of the topology of the self-motion manifold. \cite*{kreutz1990kinematic,kreutz1992kinematic}
provided a detailed analysis of the conventional SEW angle with application to 3R-R-3R arms and the RRC arm, which has no intersecting \new{consecutive }joint axes. They also provided the expression for the SEW Jacobian and singularity analysis.

\new{
Parameterizing the redundant degree of freedom is not strictly necessary to control a 7R arm. For example, one can use resolved velocity or resolved acceleration control using the weighted or unweighted Jacobian pseudoinverse, or perform optimization of some other task which is not a function of joint angles. A critical problem with these controllers is non-cyclicity \citep*{swaim1994use, stanczyk2006development}. These controllers are non-conservative, so if the robot makes a closed loop in its end effector path, the elbow may not return to its original position. This is unacceptable in any industrial or mission-critical scenario. Furthermore, these differential redundancy resolution techniques have stability problems: Minimizing joint velocity, joint torque, or joint acceleration locally does not prevent the robot from approaching a singularity \citep*{baillieul1987kinematically, hollerbach1987redundancy}. Extra work must be done to stabilize such a controller to prevent it from driving itself into a singularity \citep*{o2000instability}.
}

\new{
On the other hand, augmenting the task space with a function of the joint angles, such as the SEW angle, ensures cyclical behavior and gives direct control of the redundant degree of freedom over the entire trajectory~\citep*{seraji1989configuration}. 
By using a global rather than local approach,
    task-space behavior is immediately known without using IK.
    This leads to intuitive teleoperation and programming
    and allows for transferring task-space behavior between robots or even between humans and robots \citep*{stanczyk2006development, lamperti2015redundancy}.

It is no wonder that SEW angle is widely used for teleoperation in 
    space \citep*{tsumaki2001numerical, zhu2021design, pryor2023teleoperation},
    underwater \citep*{carignan2000partitioned, carignan2001controlling, naylor2007visual, scott2008line}, and
    in surgery \citep*{su2018safety_b, su2018safety_a, su2019improved}.

The SEW angle allows for ease of global planning \citep*{pryor2023teleoperation} since the entire solution space can be considered,
and it can be easily extended to incorporate hybrid force-motion or impedance control of both end effector motion and self-motion \citep*{su2019improved, su2018safety_b, su2018safety_a, xiong2020null}.

Additionally, the SEW angle is compatible with other optimizations to avoid obstacles, joint angle limits, joint torque limits, or singularities, or to maximize metrics like force transmission ratio \citep*{carignan2000partitioned, carignan2001controlling, naylor2007visual, scott2008line, stanczyk2004development, lee2006redundancy, su2019improved, su2018safety_b, su2018safety_a}.
}

\new{There are also applications of the SEW angle not just to typical 7-DOF robots, but also to human arms and wearable robots \citep*{
kim2011redundancy,
kim2012redundancy,
li2013rotational,
wang2013closed,
liu2017analytical,
su2018online,
wang2019kinematic,
li2022human}.}

There have been several other proposed redundancy parameterizations besides the conventional SEW angle. \cite*{yan2014analytical} attempted to address the singularity issue of the conventional SEW angle by proposing a method which picks two different reference vectors and switches between the corresponding SEW angles when one of them encounters a singularity. The resulting parameterization is non-smooth, and singularities would still be present.
\new{Furthermore, this parameterization is no longer a function of the joint angles and will therefore be noncyclical. Similar issues occur for any other method which switches between parameterizations, such as the method proposed by \cite*{an2014analytical} to switch between the SEW angle and the first joint angle.}
\cite*{shimizu2007practical} used the SEW angle where the reference plane is formed by the elbow when joint three is zero. They explained the reference plane is undefined in a shoulder or elbow singularity, so there is no benefit in terms of singularity existence as compared to the conventional SEW angle.
\new{One can also parameterize the redundant DOF by choosing one joint angle and reducing the problem to the IK of a non-redundant 6R manipulator, but this introduces singularities resulting from the equivalent 6R robot. In fact, it appears that all proposed redundancy parameterizations up to now have a bidirectional singularity. This makes the stereographic SEW angle the only parameterization proposed so far with a unidirectional singularity structure.}

\cut{One can also parameterize the redundant DOF by choosing one joint angle and reducing the problem to the IK of a non-redundant 6R manipulator, but this introduces singularities resulting from the equivalent 6R robot.}
\new{Numerous authors have found the IK solutions of 7R arms parameterized by one joint angle.}
    \cite*{xu2014analytical} found the inverse kinematics for a 2R-3R\textbar\textbar-2R arm by parameterizing the redundant degree of freedom with any joint angle.
    \cite*{jiang2013integrated} found the R-R-3R-2R arm IK solution by parameterizing joint 1.
    \cite*{an2014analytical} found the IK solution for a 2R-2R-3R arm by parameterizing joint 1.
    \cite*{tondu2006closed} found IK for a 3R-R-3R arm by parameterizing joint 1, and \cite*{pfurner2016closed} found the IK for a 3R-R-3R arm by parameterizing joint 2 or 3.
    \cite*{nammoto2012analytical} found the IK solutions for an R-R-3R-2R arm parameterized by joint 1.
    \cite*{sinha2019geometric} found a 1D search-based IK for R-2R-2R-2R arms parameterized by joint 1.

Many papers provide inverse kinematics solutions using the conventional SEW angle.
    \cite*{gong2019analytical, tian2021analytical, faria2018position} used the conventional SEW angle to provide closed-form IK for 3R-R-3R arms.
    \cite*{an2014analytical} found a closed-form IK solution for 2R-2R-3R arms.
Many papers used an iterative IK method with an approximate closed-form solution as a starting point:
    \cite*{wang2021inverse} for an R-2R-R-3R arm, and
    \cite*{ma2021precise} for a 2R-3R\textbar\textbar-2R arm. \citep*{zhao2023inverse} also found an iterative solution for an 2R-3R\textbar\textbar-2R. The paper states closed-form IK is not possible when using SEW angle.
However, \citep*{jin2020efficient} solved IK for a 2R-3R\textbar\textbar-2R manipulator in closed form by defining the SEW plane to be perpendicular to joint axes 3, 4, and 5.
\new{
Until now, no one has proposed a unified method of inverse kinematics for any 7R manipulator parameterized by SEW angle which does not rely on the Jacobian.
}

\cut{
There are also applications of the conventional SEW angle not just to typical 7-DOF robots, but also to human arms and wearable robots }\cutcitation{\citep*{
kim2011redundancy,
kim2012redundancy,
li2013rotational,
wang2013closed,
liu2017analytical,
su2018online,
wang2019kinematic,
li2022human}.}\cut{
The conventional SEW angle can also be used to help with redundancy resolution. \mbox{\cite*{lamperti2015redundancy}} used conventional SEW angle to propose a human-like redundancy resolution method.
\mbox{\cite*{xiong2020null}} used the conventional SEW angle to parameterize the redundant degree of freedom for null-space impedance control.
Algorithmic singularities remain an issue in these methods and would need to be avoided. They are not explicitly addressed in these papers.
}

\subsection{Robot Kinematics}
\begin{table}[t]
    \small\sf\centering
    \caption{Nomenclature.}
    \begin{tabular}{l @{\quad} l}
        \toprule
        \multicolumn{2}{l}{Coordinate-Free Notation}\\
        \midrule
        \(\calO\)     &  Point in Euclidean space. \\
        \(\vec p\)    &  Vector in Euclidean space. \\
        \(\calE\)     &  Orthonormal frame, \(=[ \vec e_x \  \vec e_y \ \vec e_z]\). \\
        \(\Rot(\vec h,\theta)\) & Rotation operator about \(\vec h\) over angle \(\theta\). \\
        \midrule
        \multicolumn{2}{l}{Robot Kinematics}\\
        \midrule
        \(\calO _S\) & Shoulder. \\
        \(\calO _E\) &  Elbow. \\
        \(\calO _W\) & Wrist. \\
        \(\calO _C\) & Projection of elbow on shoulder-wrist line. \\
        \(J\)   & End effector Jacobian.\\
        \(J_E\) & Elbow Jacobian. \\
        \(J_W\) & Wrist Jacobian. \\
        \midrule
        \multicolumn{2}{l}{General SEW Angle}\\
        \midrule
        \(k_{SEW}\) & \(= p_{SW}^\times p_{SE}\). \\
        \(n_{SEW}\) & \(= k_{SEW} / \norm{k_{SEW}}\). SEW plane normal. \\
        \(\psi\) & SEW angle.\\
        \(e_x {=} f_x(p_{SW})\) & Reference direction function.\\
        \((e_x,e_y, e_{SW})\) & SEW angle coordinate frame.\\
        \(J_{f_x}\) & Reference direction Jacobian w.r.t. \(p_{SW}\).\\
        \(J_\psi\) & SEW angle Jacobian. \\
        \(J_A\) & Augmented Jacobian. \\
        \midrule
        \multicolumn{2}{l}{Conventional SEW Angle}\\
        \midrule
        \(e_r\) & Arbitrary unit \underline{r}eference vector. \\
        \(k_y\) & \(= p_{SW}^\times e_r\), meaning \(e_y = k_y / \norm{k_y}\). \\
        \midrule
        \multicolumn{2}{l}{Stereographic SEW Angle}\\
        \midrule
        \(e_t\) & Unit \underline{t}ranslation vector. \\
        \(k_{rt}\) & \(= (e_{SW} - e_t)^\times e_r\).\\
        \(k_x\) & \(= k_{rt} ^\times p_{SW}\), meaning \(e_x = k_x / \norm{k_x}\). \\
        \bottomrule
    \end{tabular}

    \label{tab:nomenclature}
\end{table}
We use the coordinate-free notation shown in Table~\ref{tab:nomenclature}. Vector \(\vec p\) represented in frame \(\calE_a\) is the \(\rr 3\) vector \(p_a = \calE_a^* \vec p\), where
\begin{equation}
\calE_a^*=\begin{bmatrix} \vec e_x \cdot \\ \vec e_y \cdot \\ \vec e_z \cdot \end{bmatrix}
\end{equation}
is the adjoint of \(\calE_a\). Frame \(\calE_b\) represented in frame \(\calE_a\) is the \(SO(3)\) matrix \(R_{ab}=\calE_a^* \calE_b\). Rotation \(\Rot(\vec h,\theta)\) in frame \(\calE_a\) is the \(SO(3)\) matrix \({\tt R}(h_a,\theta)=e^{h_a^\times \theta}\), where \((\cdot)^\times\) is the \(3\times 3\) skew-symmetric matrix representation of the cross product:
\begin{equation}
    h^\times := \begin{bmatrix}  0 & -h_z & h_y \\ h_z & 0 & -h_x \\ -h_y & h_x & 0 \end{bmatrix}.
\end{equation}
For a unit vector \(h\), \(-h^{\times^2} = -h^\times h^\times = I - h h\tr\) is the projector onto the orthogonal complement of \(h\).

\begin{figure}[t]
    \centering
    \includegraphics[scale = 0.5, clip]{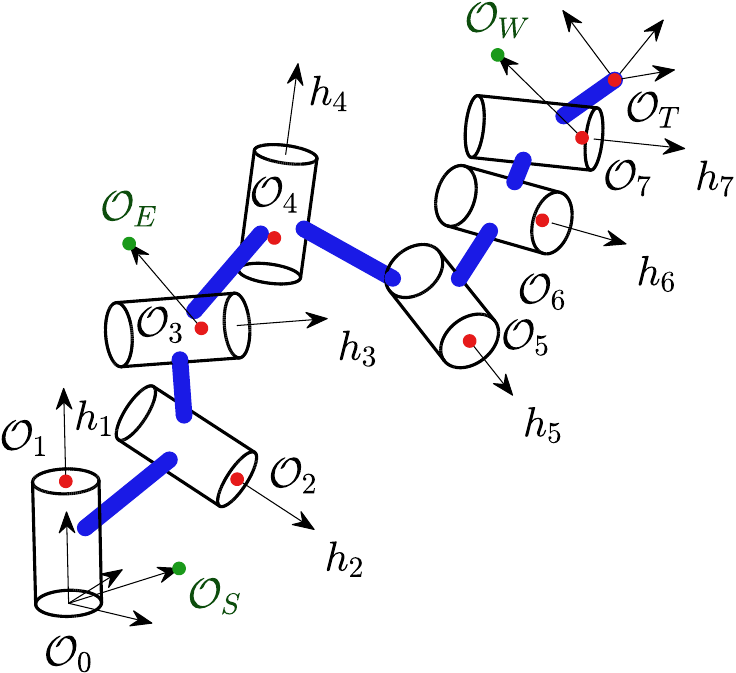}
    \caption{General 7R robot arm with example shoulder, elbow, and wrist points \(\calO_S\), \(\calO_E\), and \(\calO_W\) fixed in \(\calE_0\), \(\calE_3\), and \(\calE_7\), respectively. In general, these points may be placed anywhere in the kinematic chain.}
    \label{fig:arbitrary_7_dof}
\end{figure}

Consider a 7R robot as shown in Figure~\ref{fig:arbitrary_7_dof} with joint angles \(q = [q_1 \ q_2 \ \cdots \ q_7]\tr\).
We will use the product of exponentials approach to describe the arm kinematics.
Denote the base frame and base frame origin as \((\calE_0, \calO_0)\) and the end effector task frame and task frame origin as \((\calE_T, \calO_T)\).
Choose the link frame origins \(\calO_i\) along each unit joint axis \(\vec h_i\).
Let \(\vec p_{ij}\) be the vector from \(\calO_{i}\) to \(\calO_j\). Define the \(i\)th frame \(\calE_i=\Rot(\vec h_i, q_i)\calE_{i-1}\). 
Then the forward kinematics of the task frame represented in the base frame is
\begin{subequations}\label{eq:fwdkin}
\begin{align}
R_{0T} ={}& R_{01}R_{12}R_{23}R_{34}R_{45}R_{56}R_{67}R_{7T}\\
\begin{split}
p_{0T} ={}& p_{01}+R_{01}p_{12}+R_{02}p_{23}+R_{03}p_{34} \\
{}& + R_{04}p_{45} + R_{05} p_{56} + R_{06} p_{67} + R_{07} p_{7T}
\end{split}
\end{align}
\end{subequations}
where \(R_{ij}=R_{i,i+1}\dotsm R_{j-1,j}\), \(h_i\) and \(p_{i-1,i}\) are constant \(\rr 3\) vectors representing \(\vec h_i\) and \(\vec p_{i-1,i}\) in \(\calE_{i-1}\), \(R_{i-1,i}= e^{h_i^\times q_i}\) is \(\Rot(\vec h_i,q_i)\) represented in \(\calE_{i-1}\), \(R_{7T}\) is a constant \(SO(3)\) matrix for the fixed wrist-tool transform, and \(p_{7T}\) is the constant tool offset in the 7 frame. The constant vectors \(h_i\), \(p_{i-1,i}\), \(p_{7T}\) and constant transform \(R_{7T}\) are obtained by putting the arm in the zero configuration \(q=0\).

Define three points \(\mathcal O_S\), \(\mathcal O_E\), and \(\mathcal O_W\) to represent the robot shoulder, elbow, and wrist, respectively. The location for these points are arbitrary, but different choices will lead to different inverse kinematics methods and different singularity structures. For easier inverse kinematics, \(\mathcal O_S\) should be constant in the base frame and \(\calO_W\) should be constant in the 7 frame (which also means it is constant in the tool frame). \(\calO_E\) must be placed somewhere in the kinematic chain such that it rotates around the line passing through \(\calO_S\) and \(\calO_W\) as the robot moves through its redundant degree of freedom while keeping the end effector pose constant. For many robots, a good choice is to make \(\calO_E\) constant in frame \(\calE_3\) or \(\calE_4\). 
In this paper, we will use \(\calE_3\) for illustration. Representing these points in their respective constant frames, we have constant vectors \(p_{0S}\), \(p_{3E}\), and \(p_{7W}\).
For a given pose, we can find \(p_{SW}\) and \(p_{SE}\), which are \(\vec p_{SW}\) and \(\vec p_{SE}\) represented in \(\calE_0\). The shoulder-wrist vector is
\begin{equation}
    p_{SW} = p_{0T} - R_{07}p_{7T} + R_{07}p_{7W} - p_{0S},
    \label{eq:p_SW}
\end{equation}
and the shoulder-elbow vector is
\begin{equation}
    p_{SE} = p_{01}+R_{01}p_{12}+R_{02}p_{23}+R_{03}p_{3E} - p_{0S}.
\end{equation}
It is often helpful for the inverse kinematics solution to set the shoulder, elbow, or wrist offset vector to \(0\) so that \(\calO_S\), \(\calO_E\), or \(\calO_W\) is coincident with the link frame origin.

\begin{figure}[t]
    \centering
    \subfloat[3R elbow.]{\includegraphics[scale=0.5, clip, trim={0.1in 0.2in 0.05in 0.05in}]{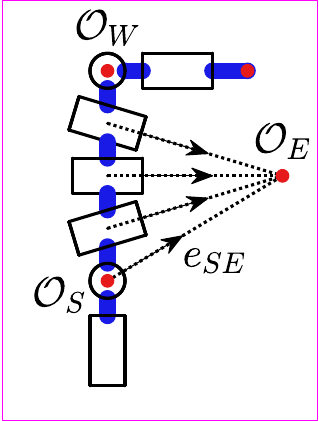}\label{fig:three_parallel_limit_midway}}
    \subfloat[3R\textbar\textbar{} elbow.]{\includegraphics[scale=0.5, clip, trim={0.1in 0.2in 0.1in 0.05in}]{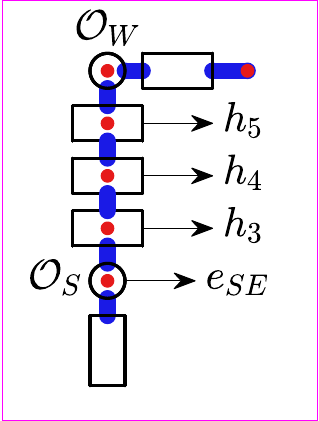}\label{fig:three_parallel_limit_limit}}
    \caption{A 3R\textbar\textbar{} joint is the limit of a 3R joint as the intersection point moves to infinity. If this joint is the elbow, then although in the limit \(p_{SE}\) has infinite length, the normalized vector \(e_{SE}\) is defined and is equal to the three parallel joint axes.}
    \label{fig:three_parallel_limit}
\end{figure}

On some robots, such as those with three consecutive parallel axes, it may be helpful to define the shoulder-elbow vector \(p_{SE}\) to be a unit vector \(e_{SE}\) which is equal to one of the joint axes represented in the base frame. For example, we may pick \(p_{SE} = e_{SE} = R_{02}h_3\). Since three parallel joint axes are the limit of three intersecting axes where the point of intersection moves infinitely far away, this can be interpreted as choosing \(p_{SE}\) to be the normalized vector pointing at this intersection (Figure~\ref{fig:three_parallel_limit}). For some robots this leads to closed-form inverse kinematics using the subproblem decomposition approach.

\section{General SEW Angle}\label{sec:gen_SEW}

\subsection{Kinematics Description}

Consider a 7-DOF arm with shoulder, elbow, and wrist defined. The SEW (shoulder-elbow-wrist) angle \(\psi\), also commonly referred to as the arm angle or swivel angle, is the angle of the shoulder-elbow vector \(p_{SE}\) about the shoulder-wrist vector \(p_{SW}\) with respect to some reference vector. Consider an arbitrary mapping, called the reference direction function, from the shoulder-wrist vector \(p_{SW}\) to a unit vector \(e_x\) that is orthogonal to \(p_{SW}\). Denote this mapping by \(e_x = f_x(p_{SW})\). From this we can form an orthonormal basis \((e_x,\ e_y,\ e_{SW})\), where \(e_{SW}\) is the normalized \(p_{SW}\) and \(e_y=e_{SW}^\times e_x\). This frame may be used to measure the SEW angle.

\begin{figure}[t]
    \centering
    \includegraphics[scale = 0.5, clip]{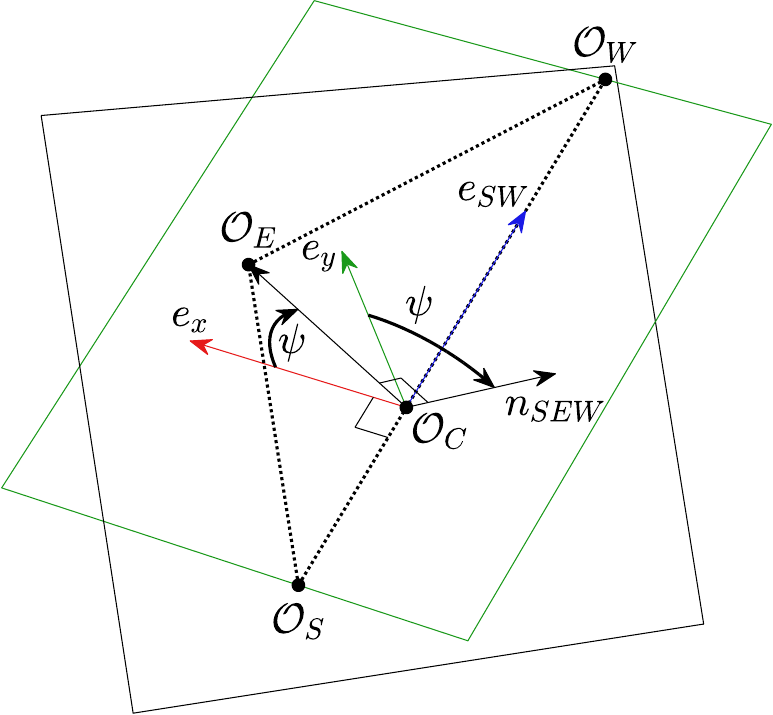}\\ \vspace{1em}
    \includegraphics[scale = 0.5, clip]{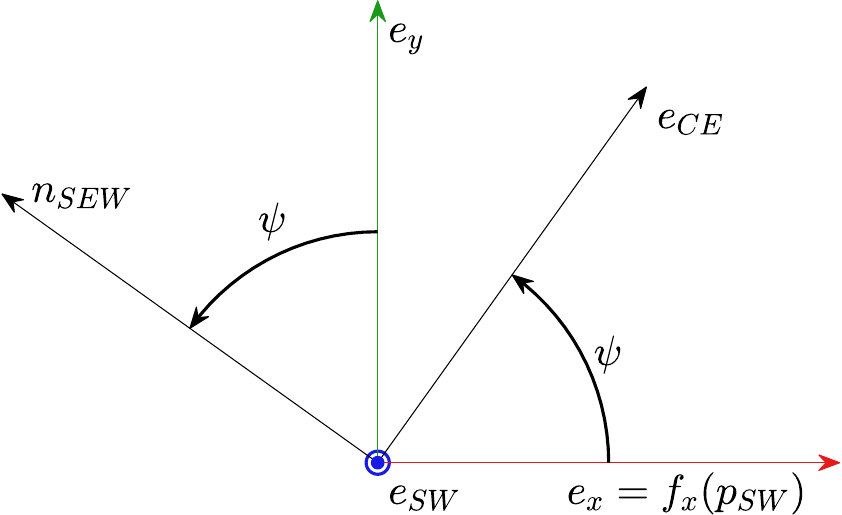}
    \caption{The SEW angle \(\psi\) is the angle of the elbow measured from \(e_x=f_x(p_{SW})\) about \(e_{SW}\), which is also the angle of the SEW plane normal vector \(n_{SEW}\) measured from \(e_y\) about \(e_{SW}\). In the general SEW angle, the reference direction function \(f_x(p_{SW})\) is arbitrary but with the constraints that the output is unit length and orthogonal to \(p_{SW}\).}
    \label{fig:axes_2d}
\end{figure}

There are two equivalent definitions of the SEW angle, as shown in Figure~\ref{fig:axes_2d}. The elbow definition is that \(\psi\) is the angle between shoulder-elbow vector \(p_{SE}\) and reference vector \(e_x\) measured along the shoulder-wrist rotational axis \(e_{SW}\). The plane definition is that \(\psi\) is the angle between \(n_{SEW}\), which is the normal vector of the SEW plane containing the shoulder, elbow and, wrist, and \(e_y\), which is the normal vector of the reference plane containing the shoulder, wrist, and reference vector \(e_x\). While the elbow definition is helpful for forward kinematics, the plane definition is more useful for inverse kinematics when using the subproblem decomposition approach.

Using the elbow definition, the SEW angle is
\begin{equation}
    \psi = \argmin_\theta \norm{ \rot(e_{SW}, \theta) e_x - p_{SE} }.
\end{equation}
We can also rewrite this more compactly by removing the component of \(p_{SE}\) along \(e_{SW}\). Let \(\calO_C\) be the point on \(p_{SW}\) such that \(p_{CE}\) is orthogonal to \(e_{SW}\), i.e.,
\begin{equation} \label{eq:p_CE}
        p_{CE} = -e_{SW}^{\times^2} p_{SE}.
\end{equation}
Then, 
\begin{equation} \label{eq:e_CE}
    e_{CE} = \rot(e_{SW},\psi) e_x,
\end{equation}
where \(e_{CE} = p_{CE} / \norm{p_{CE}}\).

Using the plane definition, the SEW angle is 
\begin{equation} \label{eq:n_SEW}
    n_{SEW} = \rot(e_{SW},\psi) e_y,
\end{equation}
where the normal of the plane containing the shoulder, elbow, and wrist is
\begin{subequations}
\begin{align}
    n_{SEW} &= \frac{k_{SEW}}{\norm{k_{SEW}}},\\ \label{eq:nsew}
    k_{SEW} &= p_{SW}^\times p_{SE}.
\end{align}
\end{subequations}
Subproblem~1 \citep*{elias2024canonical} may be used to solve the forward kinematics for the SEW angle, which is
\begin{subequations}
\begin{align}
    \psi &= \mbox{atan2}\left(
    e_y\tr p_{SE},
    e_x\tr p_{SE}
    \right)\\
    &= \mbox{atan2}\left( \label{eq:psi_p_CE}
    e_y\tr p_{CE},
    e_x\tr p_{CE}
    \right)\\
    &= \mbox{atan2}\left(
    -e_x\tr n_{SEW},
    e_y\tr  n_{SEW}
    \right)\\
    &= \mbox{atan2}\left(
    -e_x\tr k_{SEW},
    e_y\tr  k_{SEW}
    \right).
\end{align}
\end{subequations}

To perform inverse kinematics, we can use a given \((R_{0T}, p_{0T})\) to find \(p_{SW}\) from \eqref{eq:p_SW}. Next, use the reference direction function \(f_x(p_{SW})\) to find \(e_x\) or \(e_y\). Then, for a given \(\psi\), we can calculate \(e_{CE}\) or \(n_{SEW}\) using \eqref{eq:e_CE} or \eqref{eq:n_SEW}, respectively, which may then be used to determine the joint angle vector \(q\). For the subproblem decomposition approach in Section~\ref{sec:IK}, we always calculate \(n_{SEW}\).

Some robots are not easily parameterized using the SEW angle. For example, the FANUC R-1000iA/120F-7B is an R-3R\textbar\textbar-3R robot \citep*{fanuc}. The redundant degree of freedom in this family of robots is the movement of joints 2, 3, and 4, along with the corresponding movement in the spherical wrist, which results in the shortening or lengthening of the distance between joints 2 and 4 \citep*{shi2021kinematics}. However, with \(\calO_S\) placed at joint 1 and \(\calO_W\) placed at the spherical wrist, there is no movement of joints 2, 3, or 4 around \(e_{SW}\) during self-motion; they stay in the same half plane and the SEW angle would fail to parameterize the redundant degree of freedom. Robots such as this are better parameterized by specifying a joint angle. (A good choice for this robot is \(q_3\).)

To find the Jacobian of the SEW angle with respect to the joint angles, \(J_\psi\), we will write it in terms of the partial Jacobians mapping joint angular velocities to the linear velocity of \(\calO_E\) and \(\calO_W\):
\begin{equation}
\dot p_{SE}  = \dot p_{0E} = J_{E} \dot q,  \quad
\dot p_{SW}  = \dot p_{0W} = J_{W} \dot q 
\end{equation}
If \(\calO_S\) is not constant in the base frame, then \(J_S\) is nonzero and we must subtract  \(J_S \dot q\).

Taking the total derivative of \eqref{eq:psi_p_CE}, we obtain
\begin{equation} \label{eq:total_derivative}
    \dot \psi = \frac{1}{\norm{p_{CE}}} (e_{SW}^\times e_{CE})\tr \dot p_{CE} - e_y \tr \dot e_x.
\end{equation}
Geometrically, these two terms are the angular velocities of \(p_{CE}\) and \(e_x\) around the axis of rotation \(e_{SW}\). Define \(J_{f_x}\) such that \(\dot e_x = J_{f_x} \dot p_{SW}\), which depends on the choice for the function \(f_x(p_{SW})\). Then by using \eqref{eq:p_CE}, we can expand \eqref{eq:total_derivative} as
\begin{multline}
    \dot \psi =
    \frac{1}{\norm{p_{CE}}} (e_{SW}^\times e_{CE})\tr \dot p_{0E}\\
    - \left( e_y\tr J_{f_x} + \frac{e_{SW}\tr p_{SE}}{\norm{p_{SW}}\norm{p_{CE}}} (e_{SW}^\times e_{CE})\tr 
     \right) \dot p_{0W}
\end{multline}
Since we have
\begin{equation}
    \dot \psi = J_\psi \dot q
\end{equation}
we can write the SEW Jacobian as
\begin{equation}
    J_\psi = J_{\psi, E} J_E + J_{\psi, W} J_W ,
\end{equation}
where
\begin{align}
    J_{\psi, E} &= \frac{(e_{SW}^\times e_{CE})\tr}{\norm{p_{CE}}},\\
    J_{\psi, W} &= -e_y\tr J_{f_x}
  - \frac{e_{SW}\tr p_{SE}}{\norm{p_{SW}}\norm{p_{CE}}} (e_{SW}^\times e_{CE})\tr.
\end{align}
We can form the \(7 \times 7\)  augmented Jacobian \(J_A\) by stacking \(J\) and \(J_\psi\):
\begin{equation}
    J_A = \begin{bmatrix}
        J \\ J_\psi
    \end{bmatrix},\ 
    \begin{bmatrix}
        \omega \\ v \\ \dot\psi
    \end{bmatrix} = J_A \dot q.
\end{equation}
The augmented task space, which has the end effector orientation as well as the SEW angle, has 7 degrees of freedom.

\subsection{Singularity Conditions}
\begin{table*}[t]
    \small\sf\centering
    \caption{Singularity conditions. Cases 2 and 3 are both algorithmic singularities.}
    \begin{tabular}{l l l}
        \toprule
        Condition & Singularity Name & Description \\
        \midrule
        1. \(J\) loses rank                                         & Kinematic     & End effector cannot move in one direction\\
        \qquad A. Null space of \(J\) tangent to self-motion manifold & Internal    & Extra continuous self-motion possible\\
        \qquad B. Null space of \(J\) not tangent to self-motion manifold & Boundary& Self-motion is instantaneous\\
        2. \(J_A\) singular (Full rank \(J\) and \(J_\psi\))        & Augmentation  & Self-motion doesn't change SEW angle\\
        3. \(J_\psi\) undefined                                     & SEW Angle     & SEW angle undefined\\
        \qquad A. \(\calO_S\), \(\calO_E\), \(\calO_W\) collinear   & Collinear     & Depends on choice of \(\calO_S\), \(\calO_E\), \(\calO_W\) \\
        \qquad B. \(J_{f_x}\) undefined                             & Coordinate    & Depends on choice of \(f_x(p_{SW})\)\\
        \bottomrule
    \end{tabular}
    \label{tab:singularity_cases}
\end{table*}

When controlling a robot using end effector pose augmented with SEW angle, singularities occur when the matrix \(J_A^{-1}\) does not exist. When a robot is close to a singularity, there may be large internal joint motion which is undesirable and possibly dangerous\new{, as well as issues with iterative algorithm convergence and numerical precision}. There are a number of conditions which result in a singularity, as shown in Table~\ref{tab:singularity_cases}. Multiple singularity types can occur simultaneously.

The self-motion manifold for 7-DOF spatial manipulators is a curve in joint space for which all points map to the same end effector pose \citep*{burdick1989characterization}. A single end effector pose may have multiple self-motion manifolds, where each manifold belongs to a different inverse kinematics branch. This manifold exists independently of any parameterization of the redundant degree of freedom, and the goal of parameterizations such as SEW angle is to assign a value to each point on the self-motion manifold. When \(J\) is full rank, the self-motion manifold is one-dimensional and the null space of \(J\) is the tangent to the self-motion manifold.

A kinematic singularity occurs when \(J\) loses rank, meaning some spatial velocity of the end effector cannot be achieved. This is a condition that depends on the kinematics and joint angles of the robot and does not depend on the parameterization of the redundant degree of freedom. There are two types of kinematic singularities: Internal singularities and boundary singularities. At internal singularities, e.g., when two joint axes are collinear, the null space of \(J\) is tangent to the self-motion manifold. A new degree of freedom for self-motion is introduced, but parameterizations such as the SEW angle may not be able to parameterize this degree of freedom. At boundary singularities, e.g., when two links are collinear, the null space of \(J\) is not tangent to the self-motion manifold, and self-motion is only instantaneous. In some cases, the entire null space of \(J\) causes only instantaneous rather than continuous self-motion, and the self-motion manifold degenerates into a point: The self-motion manifold is zero-dimensional and any parameterization for the redundant degree of freedom is unnecessary because for each inverse kinematics branch there is only one choice for \(\psi\).

For all conditions other than the kinematic singularity, we call them algorithmic singularities as they occur not because of the robot itself but because of the algorithms we use to parameterize the redundancy. Many authors \citep*{kreutz1990kinematic, chiaverini1997singularity, marani2003algorithmic} have defined algorithmic singularities to occur when \(J_A^{-1}\) does not exist and \(J\) and \(J_\psi\) are separately full rank. However, as in \cite*{wang1995identification, faria2018position}, we generalize the definition of algorithmic singularity to include cases where \(J_\psi\) may not be full rank (i.e., may be a zero vector) or may not exist.

One type of algorithmic singularity is when \(J_A\) loses rank, which is possible even when \(J\) and \(J_\psi\) are full rank. We call this the augmentation singularity, as it occurs only when end effector and SEW angle rates are considered simultaneously. Algebraically, this singularity occurs when \(J_\psi\) is linearly dependant with the rows of \(J\). Geometrically, this means self-motion does not cause a change in SEW angle.

The second type of algorithmic singularity is when \(J_A^{-1}\) does not exist because \(J_\psi\) does not exist, which we call the SEW angle singularity. There are two cases for the SEW angle singularity. The first case, called the collinear singularity, happens when \(\calO_S\), \(\calO_E\), and \(\calO_W\) are collinear, which causes \(p_{SW} = 0\) or \(p_{CE} = 0\). The second case, called the coordinate singularity, is when \(J_\psi\) does not exist because \(J_{f_x}\) does not exist. In fact, in Section~\ref{sec:singularityexistence}, we show that for any choice of \(f_x(p_{SW})\) there will always be choices of \(p_{SW}\) that causes singularities in \(J_{f_x}\).

Given that the collinear singularity is unavoidable for some robots, it can be beneficial to place  \(\calO_S\), \(\calO_E\), and \(\calO_W\) such that the SEW angle is undefined exactly when there is only one choice for elbow position per IK branch due to boundary singularities. This reduces the total number of singularities in the workspace. In the case of a 2R-2R-3R robot, placing the shoulder, elbow, and wrist at the joint intersections means that SEW angle is undefined when the robot has no self-motion of the elbow to parameterize anyway.

The condition of \(J_\psi\) losing rank, meaning \(J_\psi =0\), may occur for a general parameterization of the redundant degree of freedom, but for most reasonable choices of placement of  \(\calO_S\), \(\calO_E\), and \(\calO_W\), this condition will not occur for the SEW angle. \cite*{choi2004multiple} called this case the secondary task singularity.

\subsection{Conventional SEW Angle}
\label{sec:classicalSEW}
\begin{figure}[t]
    \centering
    \subfloat[Conventional SEW angle.]{\includegraphics[scale = 0.5, clip]{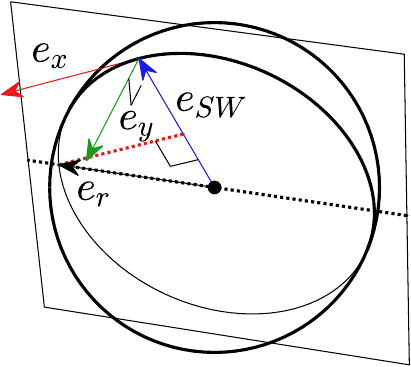}\label{fig:f_x_conv}\hspace{6pt}}%
    \subfloat[Stereographic SEW angle.]{\hspace{6pt}\includegraphics[scale = 0.5, clip]{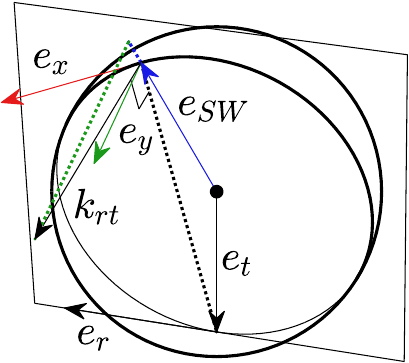}\label{fig:f_x_stereo}}
    \caption{ Geometric interpretations for \(e_x = f_x(p_{SW})\).
    (a) In the conventional SEW angle, \(e_x\) is the normalized version of the component of \(e_r\) orthogonal to \(e_{SW}\), and \(e_y\) is normal to the plane containing \(e_{SW}\) and \(e_r\).
    (b) In the stereographic SEW angle, \(k_{rt}\) is normal to the plane containing \(e_{SW}-e_t\) and \(e_r\), \(e_y\) is the normalized version of the component of \(k_{rt}\) orthogonal to \(e_{SW}\), and \(e_x\) is normal to \(k_{rt}\) and \(e_{SW}\).}
    \label{fig:f_x}
\end{figure}

The conventional SEW angle, shown in Figure~\subref*{fig:f_x_conv}, is a widely used parameterization. \cite*{kreutz1990kinematic} were the first to provide detailed analysis of the conventional SEW angle, but the idea was described earlier by \cite*{hollerbach1985optimum}.

Let \(e_r\) be an arbitrary unit reference vector. For the conventional SEW angle, we define the reference direction function \(f_x\) as
\begin{subequations}
\begin{align}
    e_x &= f_x(p_{SW}) = e_y^ \times e_{SW},\\
    e_y &= \frac{k_y}{\norm{k_y}},\\
    k_y &= p_{SW} ^ \times e_r.
\end{align}
\end{subequations}
Equivalently, \(e_x= f_x(p_{SW})\) is the normalized \(-e_{SW}^{\times^2}e_r\), so \(e_x\) is the unit vector orthogonal to \(e_{SW}\) closest to \(e_r\).

The conventional SEW angle is intuitive to understand. \(\psi = 0\) corresponds to \(e_{CE}\) pointing as much towards \(e_r\) as possible. This is similar to navigating on a globe using cardinal directions where \(e_r\) provides the north direction and the SEW angle is measured from north. 

\begin{figure}[t]
    \centering
    \subfloat[Conventional SEW angle.]{
        \begin{tabular}[b]{@{}c@{}}
             \includegraphics[scale=0.5, clip]{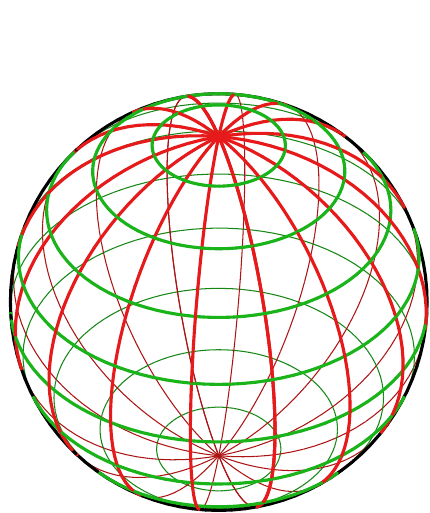}\\
             \includegraphics[scale=0.5, clip]{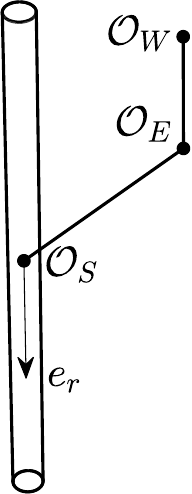}
        \end{tabular} \label{fig:singularity_locations_conv}
    }%
    \subfloat[Stereographic SEW angle.]{
        \begin{tabular}[b]{@{}c@{}}
             \includegraphics[scale=0.5, clip]{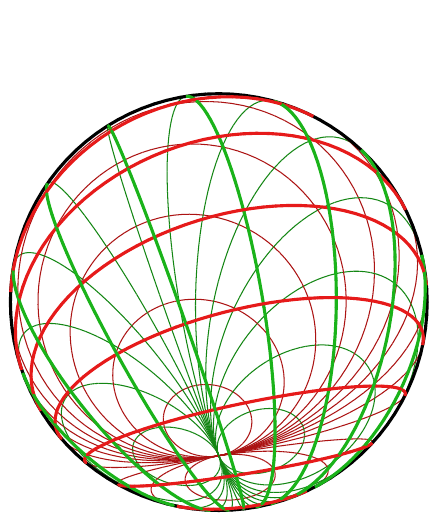}\\
             \includegraphics[scale=0.5, clip]{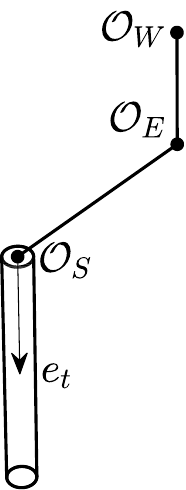}
        \end{tabular} \label{fig:singularity_locations_stereo}
    }%
    \caption{
    Grid representing \((e_x, e_y)\) for any choice of \(e_{SW}\) (top), and cylinder representing locations for \(\calO_W\) where the robot is at or near a coordinate singularity. (a) The conventional SEW angle has two singularities of order one, and the coordinate axes are lines of latitude and longitude. This results in a bidirectional singularity region in task space. (b) The stereographic SEW angle has a single singularity of order 2, which results in a unidirectional singularity region.}
    \label{fig:singularity_locations}
\end{figure}

The conventional SEW angle becomes undefined and a coordinate singularity occurs when \(k_y = p_{SW}^\times e_r =0\), which occurs when \(p_{SW}\) is collinear with \(e_r\). In this case. all possible choices of \(e_{CE}\) are equally close to \(e_r\).
This is akin to being on the north or south pole of a globe where there is no north direction and cardinal directions are undefined. If the shoulder is constant in the base frame, then this singularity occurs when the wrist is on the line spanned by \(e_r\) passing through the wrist, as shown in Figure~\subref*{fig:singularity_locations_conv}.

Using the elbow definition of the SEW angle, the
conventional SEW angle can be succinctly expressed as
\begin{equation}
    \psi = \argmin_\theta \norm{ \rot(e_{SW}, \theta) e_r - p_{SE} },
\end{equation}
(notice we may use \(e_r\) in place of \(e_x\)), which may be directly solved using Subproblem~1 as
\begin{subequations}
\begin{align}
    \psi &= \mbox{atan2}\left(
    (e_{SW} ^\times e_r)\tr p_{SE},
    -(e_{SW} ^{\times\sq} e_r)\tr p_{SE}
    \right)\\
    &= \mbox{atan2}( e_{SW}\tr e_r^\times p_{CE}, e_r\tr p_{CE} ).
\end{align}
\end{subequations}
Alternatively, using the plane definition we have
\begin{subequations}
\begin{align}
    \psi
    & = \argmin_\theta \norm{ \rot(e_{SW}, \theta) k_y - k_{SEW} }\\
    & = \mbox{atan2}( (e_{SW}^\times k_y)\tr k_{SEW},-(e_{SW}^{\times^2} k_y)\tr k_{SEW})\\
    &= \mbox{atan2}(e_{SW}\tr k_y^\times k_{SEW}, k_y \tr k_{SEW}).
\end{align}
\end{subequations}
To calculate the Jacobian, we can express \(-e_y\tr \dot e_x \) as
\begin{equation}
    \frac{e_x\tr}{\norm{k_y}} \dot k_y
    = \frac{e_x\tr}{\norm{k_y}} e_r ^\times  \dot p_{SW}
    = \frac{e_{SW}\tr e_r}{\norm{k_y}}e_y\tr  \dot p_{SW}.
\end{equation}
Therefore, the Jacobian with respect to the wrist is
\begin{equation}
    J_{\psi, W} = \frac{e_{SW}\tr e_r}{\norm{k_y}}e_y\tr
  - \frac{e_{SW}\tr p_{SE}}{\norm{p_{SW}}\norm{p_{CE}}} (e_{SW}^\times e_{CE})\tr.
\end{equation}
The Jacobian becomes undefined when \(k_y = p_{SW}^\times e_r =0\), again exhibiting the algorithmic singularity condition when \(e_r\) and \(p_{SW}\) become collinear.

\section{Existence of Singularity}
\label{sec:singularityexistence}
\cite*{hollerbach1985optimum} showed any revolute manipulator will have internal singularities associated with end effector orientation because it is always possible to find a pose with all joints lying in the same plane. \cite*{gottlieb1986robots} used topology to arrive at the same conclusion: Every smooth map \(T^n \rightarrow SO(3)\) must have singularities. Furthermore, he showed it is impossible to construct a smooth left-inverse of a mapping \(T^4 \rightarrow SO(3)\), so there is no singularity-free function to map from orientation to joint angles in a 4R joint. (However, it is possible to always find joint angles to follow a given trajectory while avoiding singularities, but this does not follow a function and so will not be cyclical.)

In this section, we will show for a 7-DOF revolute arm with a sufficiently large workspace, e.g., a 3R-R-3R robot with orthogonal consecutive joints, and a task space augmented by any parameterization of the redundant degree of freedom, e.g., the SEW angle, there will always be an algorithmic singularity.

\begin{figure}[t]
    \centering
    \includegraphics[scale = 0.5, clip]{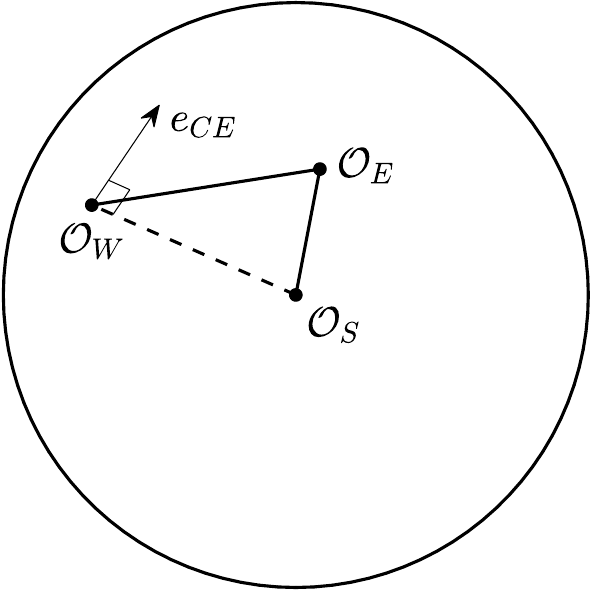}
    \caption{For each possible position of the wrist on a sphere around the shoulder while fixing the value of the arbitrary redundancy parameterization \(\phi\), define a unit vector tangent to the sphere and pointing towards the elbow. By Hairy Ball Theorem, this vector field cannot be continuous.}
    \label{fig:singularity_proof}
\end{figure}

The so-called Hairy Ball Theorem \citep*{mcgrath2016extremely} states that a continuous vector field on a sphere must have at least one point where it vanishes. We will use this theorem to show that a singularity must exist for any redundancy parameterization, \(\phi:T^7\to \rr{}\). Given any end effector pose \((R_{0T}, p_{0T})\) and \(\phi\), there exists a finite number of inverse kinematics solutions \(q\). Consider the sphere generated by a constant shoulder-wrist distance \(\norm{p_{SW}}\) such that the elbow \(\calO_E\) does not lie on the shoulder-wrist line, i.e., \(p_{SE}\) and \(p_{SW}\) are not collinear. For any point on the sphere and a constant orientation and \(\phi\), find the inverse kinematics solution \(q\) that is on the same branch of the finite number of solutions. 
Define the vector field \(e_{CE}(p_{SE})\) as the unit vector perpendicular to \(p_{SE}\) and pointing towards the elbow, which is the normalized version of \(-e_{SW}^{\times^2}p_{SE}\), as shown in Figure~\ref{fig:singularity_proof}. By the Hairy Ball Theorem, \(e_{CE}\) cannot be continuous everywhere on the sphere. This implies there is discontinuity in the joint angle while the wrist travels along the sphere, corresponding to the algorithmic singularity.

It is also straightforward to demonstrate that for the general SEW angle specifically, any choice of \(f_x(p_{SW})\) must have a singularity. If we restrict the input to have some constant nonzero  \(\norm{p_{SW}}\), then \(f_x\) defines a unit-magnitude vector field on a sphere. By the Hairy Ball Theorem, this vector field must be discontinuous. For the conventional SEW angle, the singularity occurs when \(e_{SW} = \pm e_r\). If we choose \(\phi\) to be the arm angle where \(e_x\) corresponds to \(e_{CE}\) when \(q_3 = 0\) and the robot has a spherical wrist, then the singularity occurs when \(e_{SW} = \pm h_1\).

Another common parameterization choice is \(\phi(q) = q_1\). In this case if the robot has a spherical wrist then the singularity occurs when \(e_{SW} = \pm R_{01}h_2\). 

The Poincaré–Hopf theorem \citep*{poincare1885courbes, hopf1927vektorfelder} implies that not only must there be a singularity on the sphere, but the total order of singularities must be two. For this vector field, the order of the singularity is how many times the elbow rotates about the shoulder-wrist line as the wrist travels around the singularity once. In all the examples of \(\phi\) shown above, there are two antipodal singularities of order one. 

The goal of Section~\ref{sec:stereographic_SEW_angle} is to find a redundancy parameterization that has a single singularity on this sphere of order two. This is in some sense the best-case singularity structure, as this results in a singularity along a half-line in the robot workspace. For most robots, this line can be chosen so that it goes into the structure holding the robot in place, meaning the singularity is out of the reachable workspace.

\section{Stereographic SEW Angle} \label{sec:stereographic_SEW_angle}
\subsection{Definition}
The conventional definition of the SEW angle in Section~\ref{sec:classicalSEW} encounters a singularity when the shoulder-wrist vector \(p_{SW}\) becomes collinear with the reference vector \(e_r\). As discussed in Section~\ref{sec:singularityexistence}, an algorithmic singularity is unavoidable, but we can reduce its impact by slightly changing the SEW angle definition so that the bidirectional line condition becomes a unidirectional half-line condition, as shown in Figure~\ref{fig:singularity_locations_stereo}.

For the stereographic SEW angle (Figure~\ref{fig:f_x_stereo}), we define the reference direction function \(f_x\) as
\begin{subequations}\label{eq:stereo_SEW_def}
\begin{align}
    e_x     &= f_x(p_{SW}) = \frac{k_x}{\norm{k_x}},\\
    k_x     &= k_{rt} ^\times p_{SW},\\
    k_{rt}  &= (e_{SW} - e_t)^\times e_r.
\end{align}
\end{subequations}
This also means \(e_y\) is the normalized version of \(-e_{SW}^{\times^2} k_{rt}\).
\(e_t\) is an arbitrary vector chosen to place the singularity structure. We will show we need the conditions \(\norm{e_r} = 1\), \(\norm{e_t} = 1\), and \(e_r \tr e_t = 1\) to achieve the half-line singularity condition. If we instead set \(e_t=0\) then \eqref{eq:stereo_SEW_def} becomes the conventional SEW angle definition.

Using the plane definition, the stereographic SEW angle can be written as
\begin{equation}
    \psi = \argmin_\theta \norm{ \rot(e_{SW}, \theta) k_{rt} - k_{SEW} },
\end{equation}
which may be solved using
\begin{equation}
    \psi = \mbox{atan2}(e_{SW}\tr k_{rt}^\times k_{SEW}, k_{SEW}\tr k_{rt}).
\end{equation}
Using the elbow definition, we may equivalently write
\begin{equation}
    \psi = \mbox{atan2}(k_{rt} \tr p_{CE}, - e_{SW}\tr k_{rt}^\times p_{CE}).
\end{equation}

For any \((e_r, e_t)\) with nonzero \(e_r\) we can always pick a new \((\tilde e_r, \tilde e_t)\) such that  \(e_r \tr e_t = 1\) and \(\norm{e_r} = 1\) where the direction of \(k_{rt}\) is identical according to
\begin{equation}
    \tilde e_r = \frac{e_r}{\norm{e_r}},\quad
    \tilde e_t = -\tilde e_r ^{\times^2} e_t.
\end{equation}
We can show that once we require \(\norm{e_r} = 1\) and \(e_r \tr e_t = 0\) (without loss of generality), the half-line singularity condition occurs if and only if \(\norm{e_t} = 1\).
The algorithmic singularity corresponds to \(k_x=0\), which occurs when \(k_{rt}\) is a zero vector or collinear with \(e_{SW}\).
We consider two cases below:
\begin{enumerate}
    \item \(\norm{e_t} \leq 1\): For two choices of \(e_{SW}\), \(k_{rt}=0\):
    \begin{equation} \label{eq:norm_e_SW_geq_1}
        e_{SW} = e_t \pm \sqrt{1-\norm{e_t}^2} e_r
    \end{equation} 
    This corresponds to the points on the unit sphere which intersect with the line spanned by \(e_r\) and translated by \(e_t\).
    \item \(\norm{e_t} \geq 1\): For two choices of \(e_{SW}\), \(k_{rt}\) is collinear with \(e_{SW}\):
    \begin{equation} \label{eq:norm_e_SW_leq_1}
        e_{SW} = \frac{e_t}{\norm{e_t}^2}
                 \pm \sqrt{{\norm{e_t}^2}-1}\frac{e_t ^\times e_r}{\norm{e_t}^2}.
    \end{equation} 
    This corresponds to the points on the unit sphere which is tangent to a plane passing through the line spanned by \(e_r\) and translated by \(e_t\).
    
\end{enumerate}
There are two unique singularities if \(\norm{e_t} < 1\) and 2 unique singularities if \(\norm{e_t} > 1\). By choosing \(\norm{e_t} = 1\), both \eqref{eq:norm_e_SW_geq_1} and \eqref{eq:norm_e_SW_leq_1} simplify to the half-line condition \(e_{SW} = e_t\). Geometrically, this is because the line spanned by \(e_r\) and translated by \(e_t\) only passes through one point on the unit sphere (at \(e_t\)), and the only plane tangent to the unit sphere and passing through the line spanned by \(e_r\) and translated by \(e_t\) is tangent at \(e_t\). If we instead set \(e_t=0\), then we recover the conventional SEW angle.

A good choice of \(e_t\) is to point in the opposite direction of the robot workspace (e.g., into the ground) so that the singularity will not affect the robot operation.

For the Jacobian, we must calculate
\begin{equation}
    -e_y\tr \dot e_x = -\frac{1}{\norm{k_x}} e_y \tr \dot k_x.
\end{equation}
Recall that \(k_x = k_{rt}^\times p_{SW}\), which means the derivative is
\begin{gather}
    \dot k_x =k_{rt} ^\times \dot p_{SW} - p_{SW}^\times \dot k_{rt},
\end{gather}
and so
\begin{equation}
    -e_y\tr \dot e_x = - \frac
        {(e_y^\times k_{rt})\tr}
        {\norm{ k_x }}
        \dot p_{SW}
    + \frac
        {(e_y^\times p_{SW}) \tr}
        {\norm{ k_x }}
        \dot k_{rt}.
\end{equation}
We can rewrite the first term as 
\begin{equation}
\begin{split}
    -\frac
        {(e_y^\times k_{rt})\tr}
        {\norm{ k_x }}
        \dot p_{SW}
    &= -\frac
        {e_{SW}\tr k_{rt}}
        {\norm{ k_x }}
        e_x \tr  J^W \dot q \\
    &= \frac
        {e_{SW} \tr e_t^\times e_r}
        {\norm{ k_x }}
        e_x\tr  J^W \dot q.
\end{split}
\end{equation}
Notice that
\begin{equation}
    \dot e_{SW} = -\frac{e_{SW} ^{\times^2}}{\norm{p_{SW}}} \dot p_{SW}.
\end{equation}
The second term can be written as
\begin{equation}
\begin{split}
    \frac
        {(e_y^\times p_{SW}) \tr}
        {\norm{ k_x }}
        \dot k_{rt}
    &=  \frac
        {\norm{ p_{SW} }}
        {\norm{ k_x }}
        e_x \tr
     \left(
        e_r^\times \frac
            {{e_{SW}^\times}^2}
            {\norm{ p_{SW} }}
        J_W \dot q
    \right) \\
    &= \frac
        {e_{SW} \tr e_r}
        {\norm{ k_x }}
    e_y \tr  J_W \dot q.
\end{split}
\end{equation}
This means the Jacobian with respect to the wrist is
\begin{equation}
    \begin{split}
    J_{\psi, W} ={}&     \frac{e_{SW}\tr e_r}{\norm{k_x}}e_y\tr
  + \frac{e_{SW}\tr e_t^\times e_r}{\norm{k_x}}e_x\tr \\
  &{}- \frac{e_{SW}\tr p_{SE}}{\norm{p_{SW}}\norm{p_{CE}}} (e_{SW}^\times e_{CE})\tr
  \end{split}
\end{equation}
As expected based on the previous singularity analysis, other than the \(p_{SW} = 0\) and \(p_{CE} = 0\) conditions from the general SEW angle, \(J_\psi\) is only undefined when \(k_x = 0\).

In deriving the Jacobian, we made no assumptions on the norms or orthogonality of \(e_r\) or \(e_t\). Setting \(e_t = 0\) lets us recover the Jacobian for the conventional SEW angle.

\subsection{Relationship to Stereographic Projection}
The stereographic SEW angle gets its name from stereographic projection, which is a type of one-to-one mapping between points on a sphere and a plane \citep*{needham1997visual}. Stereographic projection has the following geometric definition: Pick a point on the sphere as the pole of the projection and place the plane tangent to the sphere at the antipodal point.
Corresponding points on the sphere and plane are collinear with the pole.

\begin{figure}[t]
    \centering
    \includegraphics[scale = 0.5, clip]{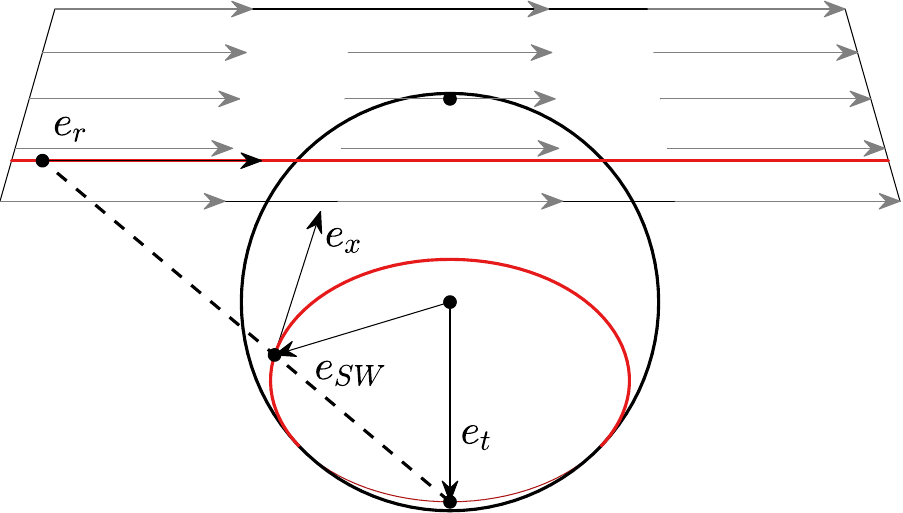}
    \caption{The stereographic projection of \(e_r\) onto \(e_{SW}\) is \(e_x\), where the pole of the projection is at \(e_t\).}
    \label{fig:stereo_proj}
\end{figure}

We can show the reference direction \(e_x\) for the stereographic SEW angle is generated by performing a stereographic projection of a constant vector field onto a unit sphere (Figure~\ref{fig:stereo_proj}). For any choice of \(e_t\) and \(e_r\), we can generate a constant vector field in the \(e_r\) direction on the projection plane and place the projection pole at \(e_t\). Then, for any choice of \(e_{SW}\), we can form a projection line passing through \(e_t\) and \(e_{SW}\) and intersecting with the projection plane. The projection of \(e_r\) onto the sphere is then the vector which is tangent to the circle formed by the intersection of the unit sphere and the plane which contains the tips of \(e_t\) and \(e_{SW}\) and is parallel to \(e_r\). This plane is normal to \(k_{rt}\), and so the projection of \(e_r\) from the plane to the sphere is \(e_x\).

\subsection{Preservation of Angles} \label{sec:preservation_of_angles}
Although it appears that we have three DOF in picking the parameters for the stereographic SEW angle (two DOF for \(e_t\) and one remaining DOF for \(e_r\)), we can show that the one DOF for \(e_r\) does not change the behavior of the SEW angle; it only changes the angle by a constant. Changing the angle of \(e_r\) around \(e_t\) is equivalent to changing the SEW angle \(\psi\) by the same angle.

Since stereographic projection is a conformal mapping, changes of angles in the projection plane get preserved after projecting onto the unit sphere. This means changing the angle of \(e_r\) on the projection plane corresponds to an equal change of angle to \(e_x\) and \(e_y\).

\begin{figure}[t]
    \centering
    \includegraphics[scale = 0.5, clip]{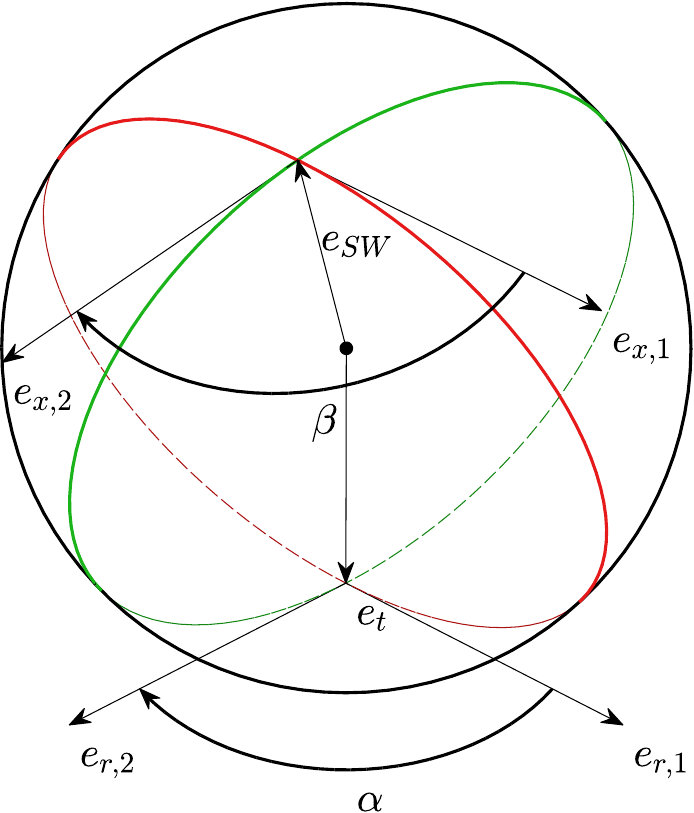}
    \caption{\(e_{x,i}\) is tangent to the circle generated from the plane through \(e_t\), \(e_t + e_{r,i}\) and \(e_{SW}\). The angles of intersections of any two circles on a sphere are equal. Therefore, \(\alpha = \beta\). }
    \label{fig:circles_proof}
\end{figure}

We can also prove this preservation of angles property directly (Figure~\ref{fig:circles_proof}). First, set \(e_r = e_{r,1}\), which results in \(e_{x,1}\). This vector is tangent to the circle which is tangent to \(e_{r,1}\) and is passing through \(e_{SW}\) and \(e_t\). Now, pick a new \(e_r = e_{r,2}\) which is \(e_{r,1}\) rotated by an angle \(\alpha\) about \(-e_t\). This generates \(e_{x,2}\) which is tangent to the circle passing through \(e_{SW}\) and \(e_t\) but tangent to \(e_{r,2}\). The angle from \(e_{x,1}\) to \(e_{x,2}\) about \(e_{SW}\) is an angle \(\beta\). The angles of intersection of any two circles on a sphere are equal, so \(\alpha = \beta\).

The preservation of angles property can aid in analyzing the behavior of SEW angle when only \(q_1\) changes in the typical case of \(e_t = -h_1\), which occurs when the first joint of the robot is pointing up and the singularity direction is chosen to point down. In this case, when \(q_1\) changes but all other joint angles are held constant, \(\psi\) changes at the same rate. This means the first element of  \(J_\psi\) is always 1.

We can compare this to the conventional SEW angle where \(e_r = h_1\). In this case, the SEW angle does not change when only \(q_1\) changes. (The conventional SEW angle has rotational invariance about \(e_r\).) This means the first element of \(J_\psi\) is always 0.

\subsection{Singularity Behavior}

The singularity for the stereographic SEW angle is of order two. This means that if \(e_{SW}\) travels in a small circle around \(e_t\), \(e_x\) rotates twice for each rotation of \(e_{SW}\). This is unlike the conventional SEW angle, where each singularity is of order one.

Although a singularity of order two would result in larger elbow motion than a singularity of order one for a fixed SEW angle as \(e_{SW}\) passes close to \(e_t\), large motion disappears if \(e_{SW}\) passes directly through \(e_t\) in a smooth path. As a path gets closer to the singularity, \(e_x\) gets closer to making a full revolution of \(2\pi\) radians. At the limit, the elbow rotates by exactly \(2\pi\), which is equivalent to not rotating at all. We can compare this to the conventional SEW angle, where passing through the singularity in a smooth path causes \(e_x\) to rotate by \(\pi\) radians.

\subsection{\new{Comparison Demonstration}\cut{Example}}

\begin{figure}[t]
    \centering
    \newoutline{\includegraphics[scale=0.5, clip]{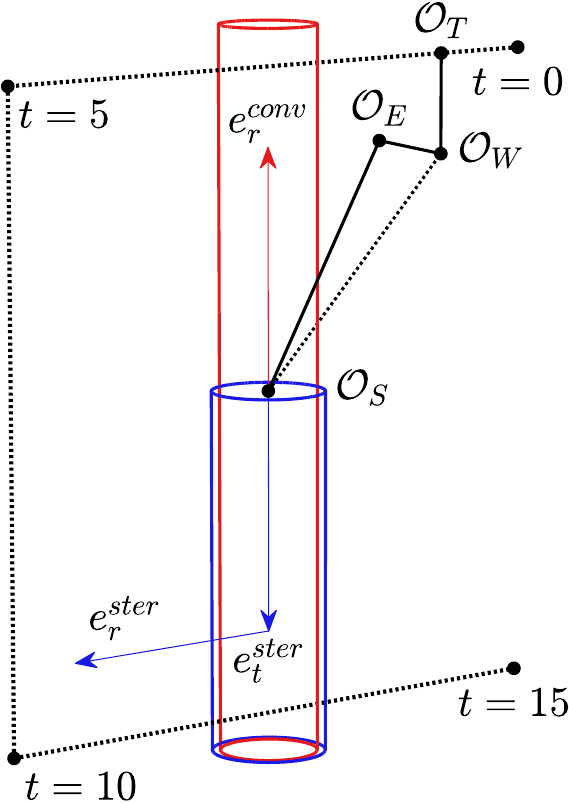}}
    \caption{\new{
    Task-space trajectory for the comparison demonstration. The end effector position follows three line segments while the end effector rotation and SEW angle are constant.
    The red and blue cylinders indicate the conventional and stereographic SEW angle singularity directions, respectively.}}
    \label{fig:kuka_unit_vectors}
\end{figure}

\begin{figure}[t]
    \centering
    \includegraphics[]{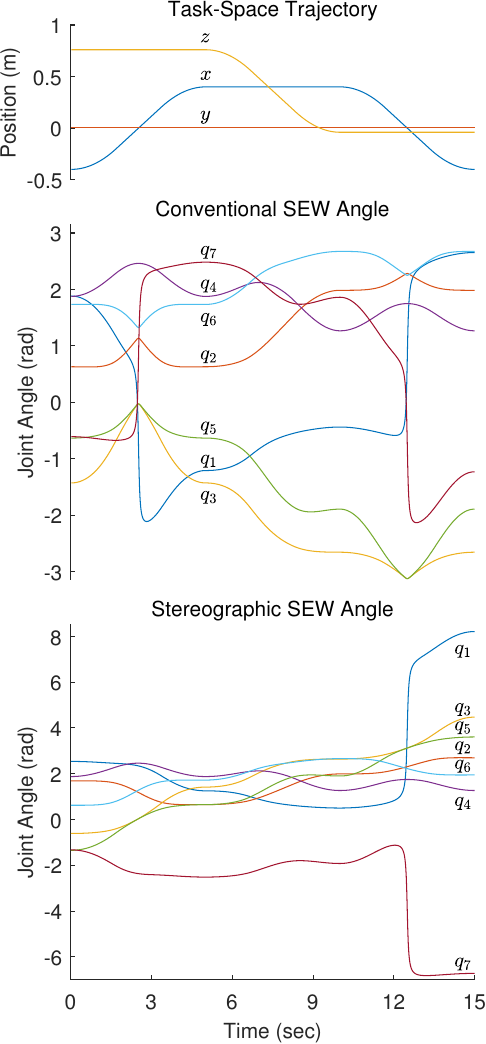}
    \caption{Joint-space trajectories using constant conventional or stereographic SEW angles. Whereas the conventional SEW angle results in large joint motion when the wrist passes near a line, the stereographic SEW angle results in large joint motion when passing near a half-line. \new{This unidirectional singularity region occurs below the robot \((z<0)\) and cannot typically be reached. (See also Extension~1.)}}
    \label{fig:demo}
    \vspace{-1em} 
\end{figure}

Two joint trajectories were generated for a KUKA LBR iiwa 14 R820 robot \citep*{KUKA} with constants \(R_{07} = I\) and \(\psi=\pi/4\): One using the conventional SEW angle with  \(e_r = [
       0 \   0 \   1
]\tr\), and the other using the stereographic SEW angle with \(e_t = [
       0 \   0 \ {-1}
]\tr\) and \(e_r = [
       0 \   1 \   0
]\tr\). The task-space trajectory for \(p_{0T}\) took a straight-line path through 4 points, with \(x=\pm 0.40\) m, \(y = 0.01\) m, and \(z = 0.36 \pm 0.40\) m.
There was a small offset in the \(y\) direction as otherwise there would not be any large joint motion for the stereographic SEW angle. For each trajectory, the initial pose was picked such that \(q_2>0\), \(q_4>0\), and \(q_6>0\). \new{The task-space trajectory is shown in Figure~\ref{fig:kuka_unit_vectors}, and the results}\cut{Results} are shown in Figure~\ref{fig:demo} \new{(Extension~1)}.

For the conventional SEW angle, the singularities occur near \(e_{SW} = \pm [
       0 \   0 \   1
]\tr\). In following the trajectory with the conventional SEW angle, there were large joint motions in two places, each corresponding to the positive and negative direction.

For the stereographic SEW angle, the singularity only occurs near \(e_{SW} = [
       0 \   0 \   {-1}
]\tr\). We see that the large joint motion only occurred at one time in the trajectory. The joint motion was also larger than in the conventional SEW case, as the singularity is of order two instead of one. The singularity occurs when the wrist is below the base, which in many cases would be out of the feasible workspace.
\section{Inverse Kinematics} \label{sec:IK}
\begin{table*}[t]
    \small\sf\centering
    \caption{Kinematic families of robots with IK solutions demonstrated in this paper. The Motoman SIA50D appears as two examples since one robot may be part of multiple kinematic families.}
    \begin{tabular}{l r@{\ }l l}
        \toprule
        IK Type & \multicolumn{2}{l}{Kinematic Family} & Robot Example \\
        \midrule Closed-Form
        & 1. & 2R-2R-3R              & FREND \citep*{debus2009overview}\\
        &    &                       & KUKA LBR iiwa 14 R820 \citep*{KUKA}\\
        & 2. & 3R-R-3R               & Motoman SIA5D \citep*{moto5d} \\
        &3. & R-R-3R\textsuperscript{E}-2R             & Motoman SIA50D  \citep*{moto50d}\\
        &4. & 2R-3R-2R  &\\
        &5. & 2R-3R\textbar\textbar-2R & SSRMS \citep*{crane1991kinematic} \\
        &&                       & SPDM \citep*{mukherji2001special}\\
        &&                       & ERA \citep*{boumans1998european}\\
        &6. & R-R-3R\textbar\textbar\textsuperscript{E}-2R & \\
        \midrule
        1D Search
        & 7. & R-2R-2R\textsuperscript{E}-2R            & Sawyer \citep*{sawyer} \\
        &&                       & Baxter \citep*{baxter}\\
        &&                       & OSAM-2 \citep*{osam2}\\
        &&                       & OB7 \citep*{OB7} \\
        &8. & 3R-R\textsuperscript{E}-2R-R             & Franka Production 3 \citep*{franka} \\
        &&                       & xArm7 \citep*{xarm}\\
        &9. & R-2R\textsuperscript{S}-R-3R             & Motoman SIA50D  \citep*{moto50d}\\
        \midrule
        2D Search
        &10. & General 7-DOF         & ABB Yumi \citep*{ABB_YUMI} \\
        &&                       & RRC \citep*{RRC}\\
        \bottomrule
    \end{tabular}

    \label{tab:robot_IK_types}
\end{table*}

\subsection{Problem Formulation}
The inverse kinematics problem for a 7R robot arm is to find all possible joint angles \(q\) corresponding to an end effector pose and SEW angle \((R_{0T}, p_{0T}, \psi)\) given the robot kinematic parameters \(\left(\{p_{i-1,i}\}_{i=1}^7,\ p_{7T},\ \{h_i\}_{i=0}^7,\ R_{7T}\right)\) and stereographic SEW parameters \(\left(e_r,\ e_t,\ p_{iS},\ p_{jE},\ p_{kW}\right)\). These inverse kinematics solutions are agnostic to the SEW formulation, and so the conventional SEW angle definition may be used instead.

The inverse kinematics procedures can apply not just to 7-DOF manipulators, but also to 6-DOF manipulators which have been provided an extra degree of freedom, say, by being placed on an omnidirectional mobile base with the location of the origin of the base specified. For example, if we use a UR5 robot \citep*{UR5}, tilt the robot so the first joint is not vertical, and pick the origin of the mobile base to be directly under the intersection of joints 1 and 2, then the system becomes a 3R-R-2R-R or 3R-2R\textbar\textbar-2R robot.

To find the inverse kinematics of a 7-DOF robot where the redundant degree of freedom is parameterized by some joint angle \(q_i\), find the 6-DOF robot generated by fixing \(q_i\) and refer to the inverse kinematics procedures provided in \cite*{elias2024canonical}.

Without loss of generality, assume \(p_{01} = 0\). (Otherwise, subtract \(p_{01}\) from \(p_{0T}\) and \(p_{0S}\).)  Rewrite the kinematics equations in terms of \(R_{07}\) and \(p_{07}\), which can be immediately calculated:
\begin{subequations} 
    \begin{align}
    R_{07} ={}& R_{0T}R_{7T}\tr = R_{01} R_{12} R_{23} R_{34} R_{45} R_{56} R_{67},\\
    \begin{split}
    p_{07} ={}& p_{0T}-R_{07}p_{7T}\\
    ={}& R_{01} p_{12} + R_{02} p_{23} + R_{03} p_{34}\\
    &+R_{04} p_{45}+R_{05} p_{56}+R_{06} p_{67}.
    \end{split}
    \end{align}
    \label{eq:fwdkin_simplified}
\end{subequations}
The inverse kinematics procedures can now be written in terms of \((R_{07},\ p_{07},\ \psi)\).
If the shoulder is constant in the 0 frame and the wrist is constant in the 7 frame, then we can use \eqref{eq:n_SEW} to find \(n_{SEW}\) since \(\psi\) is given \(p_{SW}\) is known:
\begin{equation}
    p_{SW} = p_{07} + R_{07}p_{7W} - p_{0S}.
\end{equation}
A key step in the inverse kinematics procedure is to find the elbow location, i.e., \(p_{SE}\). This sometimes involves finding the shoulder angle \(\theta_S\) or the wrist angle \(\theta_W\), defined as 
\begin{equation}
    p_{SE} = \rot(n_{SEW},\theta_S)e_{SW}\norm{p_{SE}},
\end{equation}
\begin{equation}
    p_{EW} = \rot(n_{SEW}, \theta_W) e_{SW}\norm{p_{EW}}.
\end{equation}
Note that for a given shoulder position, wrist position, and SEW angle, the elbow must be somewhere in a half plane: Given \(n_{SEW}\) and \(e_{CE}\), we require \(n_{SEW}\tr p_{SE} = 0\) and \(e_{CE}\tr p_{SE} > 0\).
To place the elbow in the correct half plane, \(\theta_S \in [0,\pi]\) and \(\theta_W \in [-\pi,0]\). A singularity occurs when \(\theta_S = 0\) or \(\theta_W = 0\), as this corresponds to the shoulder, elbow, and wrist being collinear.

\begin{figure}[t]
    \captionsetup[subfloat]{labelformat=empty}
    \centering
    \subfloat[1. 2R-2R-3R.]{\includegraphics[scale=0.5, clip, trim={0.1in 0.2in 0.1in 0.05in}]{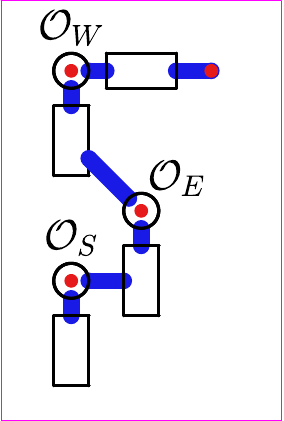}}
    \subfloat[2. 3R-R-3R.]{\includegraphics[scale=0.5, clip, trim={0.1in 0.2in 0.1in 0.05in}]{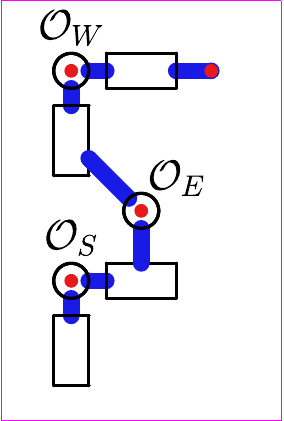}}
    \subfloat[3. R-R-3R\textsuperscript{E}-2R.]{\includegraphics[scale=0.5, clip, trim={0.1in 0.2in 0.1in 0.05in}]{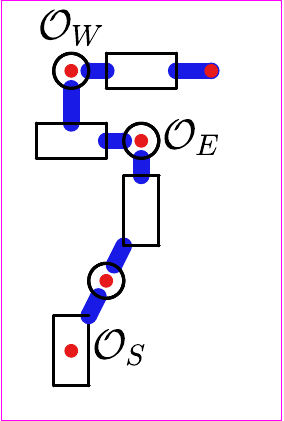}}\\
    \subfloat[4. 2R-3R-2R.]{\includegraphics[scale=0.5, clip, trim={0.1in 0.2in 0.1in 0.05in}]{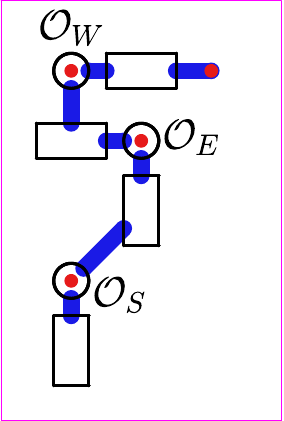}}
    \subfloat[5. 2R-3R\textbar\textbar-2R.]{\includegraphics[scale=0.5, clip, trim={0.1in 0.2in 0.1in 0.05in}]{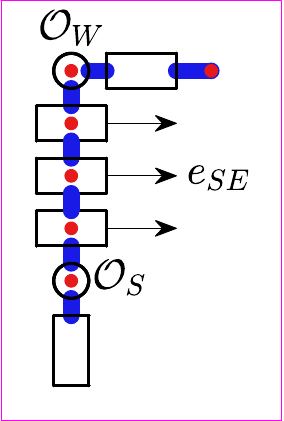}}
    \subfloat[6. R-R-3R\textbar\textbar\textsuperscript{E}-2R.]{\includegraphics[scale=0.5, clip, trim={0.1in 0.2in 0.1in 0.05in}]{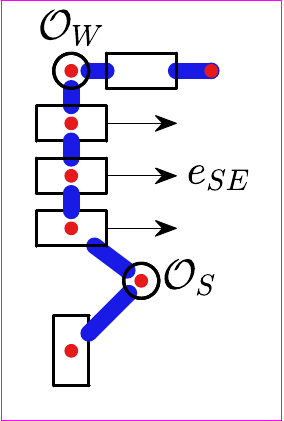}}\\
    \subfloat[7. R-2R-2R\textsuperscript{E}-2R.]{\includegraphics[scale=0.5, clip, trim={0.1in 0.2in 0.1in 0.05in}]{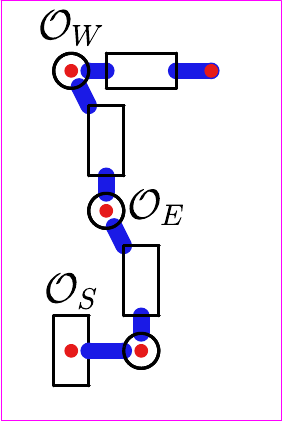}}
    \subfloat[8. 3R-R\textsuperscript{E}-2R-R.]{\includegraphics[scale=0.5, clip, trim={0.1in 0.2in 0.1in 0.05in}]{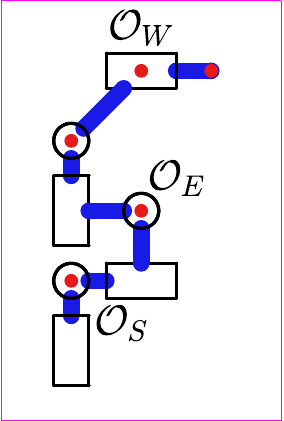}}
    \subfloat[9. R-2R\textsuperscript{S}-R-3R.]{\includegraphics[scale=0.5, clip, trim={0.1in 0.2in 0.1in 0.05in}]{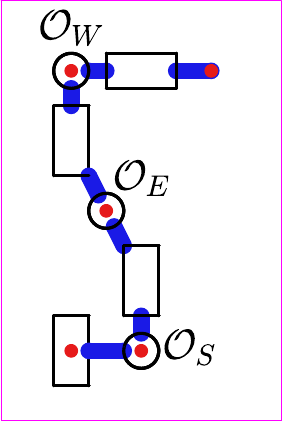}}
    \caption{
    Examples of robots in kinematic families with closed-form and 1D search solutions.}
    \label{fig:robots}
    \captionsetup[subfloat]{labelformat=parens}
\end{figure}

A number of inverse kinematics solutions are provided below, and the different kinematic families, along with examples of specific robots for some types, are shown in Table~\ref{tab:robot_IK_types} and Figure~\ref{fig:robots}. There exist many other kinematic families, but we include enough examples to demonstrate the technique of applying the subproblem decomposition method to solve the inverse kinematics problem. The reader is referred to the discussions in \cite*{elias2024canonical} for information regarding handling multiple branches, extraneous solutions, least-squares solutions, 1D and 2D searches, and internal and boundary singularities.

Closed-form solutions may exist when a robot has 3R or 3R\textbar\textbar{} joints, as a 3R joint may result in decoupling between position and orientation, and a 3R\textbar\textbar{} joint is the limit of a 3R joint as the point of intersection moves infinitely far away. (It only makes sense to place \(\calO_E\) infinitely far away, as placing \(\calO_S\) or \(\calO_W\) infinitely far away results in \(e_{SW}\) being a constant vector in the base or end effector frame.) In other cases, the solution requires a 1D or 2D search over \(q_i\), \(\theta_S\), or \(\theta_W\). \new{We also demonstrate converting a search-based solution into a polynomial root finding problem using the tangent half-angle substitution in Section~\ref{sec:polynomial_method}.}

A common step in the IK solutions below is to solve the orientation of a 3R joint. To solve such a spherical joint (shown here at the robot wrist)
\begin{equation}
    R_{45} R_{56} R_{67} = R_{47},
\end{equation}
first solve for \((q_5, q_6)\) using Subproblem~2
\begin{equation}
    R_{56} h_7 = R_{54} R_{47} h_7.
\end{equation}
Then, solve for \(q_7\) using Subproblem~1:
\begin{equation}
    R_{67} p = R_{65}R_{54} R_{47} p,
\end{equation}
where \(p\) is any vector not collinear with \(h_7\).

In the following IK solutions, always pick the origins of intersecting axes for that kinematic family to be coincident. For example, in a 3R-R-3R robot, pick  \(\mathcal O_1 = \mathcal O_2 = \mathcal O_3 = \mathcal O_S\), which means \(p_{12}=p_{23}=p_{0S}=0\). When an elbow is made of parallel joints (say, that include joint 3), pick \(p_{SE} = e_{SE} = R_{02}h_3\).

\subsection{IK Solutions}
\subsubsection{2R-2R-3R Arm}
The position equation becomes
\begin{equation}
    p_{07} = p_{SW} = R_{02} (p_{23} + R_{24}p_{45}),
\end{equation}
and given \(\theta_S\) we have
\begin{equation}
    R_{02}p_{23} = \rot(n_{SEW},\theta_S)e_{SW}\norm{p_{23}}.
\end{equation}
Use Subproblem~3 to find \(\theta_S \in [0,\pi]\):
\begin{equation}
    \norm{ \rot(n_{SEW},\theta_S)e_{SW}\norm{p_{23}} - p_{07}} = \norm{p_{45}}.
\end{equation}
Find \((q_1, q_2)\) using Subproblem~2:
\begin{equation} \label{eq:q_12_using_sp2}
    R_{12} p_{23} = R_{10} \rot(n_{SEW},\theta_S)e_{SW}\norm{p_{23}}.
\end{equation}
Similarly, find \((q_3, q_4)\) using Subproblem~2:
\begin{equation}
    R_{32}(R_{21}R_{10}p_{07}-p_{24}) = R_{34}p_{45}.
\end{equation}
Finally, find \((q_5, q_6, q_7)\) by solving the spherical wrist:
\begin{equation}\label{eq:spherical_wrist_with_RHS}
    R_{45}R_{56}R_{67} = (R_{01}R_{12}R_{23}R_{34})\tr R_{07}.
\end{equation}

\subsubsection{3R-R-3R Arm}
The position equation becomes
\begin{equation}
    p_{07} = p_{SW} =  R_{03} (p_{34} + R_{34}p_{45}).
\end{equation}
Solve for up to two solutions of \(q_4\) using Subproblem~3:
\begin{equation}
    \norm{R_{34}p_{45} + p_{34}} = \norm{p_{07}}.
\end{equation}
Represent \(R_{03}\) as three consecutive orthogonal rotations:
\begin{equation}
    R_{03} = \rot(e_{SW},\theta_a)\rot(n_{SEW},\theta_b)\rot(e_{SW},\theta_c).
\end{equation}
There are up to two solutions for \((\theta_a, \theta_b, \theta_c)\), but they represent the same \(R_{03}\), so only keep one solution. Solve for \((\theta_b, \theta_c)\) using Subproblem~2:
\begin{equation}
    \rot(n_{SEW}, \theta_b)\tr p_{07} =  \rot(e_{SW},\theta_c)(p_{34} + R_{34}p_{45}).
\end{equation}
Then, use Subproblem~4 to find \(\theta_a\), keeping only the solutions that place the elbow in the correct half plane:
\begin{equation}
    n_{SEW}\tr \rot(e_{SW},\theta_a)\rot(n_{SEW},\theta_b)\rot(e_{SW},\theta_c) p_{3E} = 0.
\end{equation}
Find \((q_1, q_2, q_3)\) by solving the spherical shoulder:
\begin{equation}
    R_{01} R_{12} R_{23} = \rot(e_{SW},\theta_a)\rot(n_{SEW},\theta_b)\rot(e_{SW},\theta_c).
\end{equation}
Similarly, find \((q_5, q_6, q_7)\) by solving \eqref{eq:spherical_wrist_with_RHS}.

\subsubsection{\texorpdfstring{R-R-3R\textsuperscript{E}-2R Arm}{R-R-3Rᴱ-2R Arm}}
Write \(\theta_W\) as
\begin{equation}
    R_{05}p_{56} = \rot(n_{SEW}, \theta_W) e_{SW}\norm{p_{56}}.
\end{equation}
Using Subproblem~5, we can find up to four solutions of \((q_1, q_2,\theta_W)\), where \(\theta_W \in [-\pi, 0]\):
\begin{multline}
    p_{07} - \rot(n_{SEW}, \theta_W) e_{SW}\norm{p_{56}} \\= R_{01} (p_{12} + R_{12} p_{23}).
\end{multline}
Next, find up to two solutions of \((q_6, q_7)\) using Subproblem~2:
\begin{equation}
    R_{67} R_{07}\tr \rot(n_{SEW}, \theta_W) e_{SW}\norm{p_{56}} = R_{65} p_{56}.
\end{equation}
To solve for \((q_3, q_4, q_5)\), solve the spherical elbow:
\begin{equation}\label{eq:spherical_elbow}
    R_{23}R_{34}R_{45} = (R_{01} R_{12})\tr R_{07} (R_{56} R_{67})\tr.
\end{equation}

\subsubsection{2R-3R-2R Arm}
Find \(\theta_S\in [0,\pi]\) using Subproblem~3:
\begin{equation}
    \norm{ \rot(n_{SEW},\theta_S)e_{SW}\norm{p_{23}} - p_{07}} = \norm{p_{56}}.
\end{equation}
Find \((q_1,q_2)\) with Subproblem~2 to solve \eqref{eq:q_12_using_sp2}. Similarly, find \((q_6, q_7)\):
\begin{equation}
    R_{67} R_{07}\tr (p_{07}- R_{02}p_{23}) = R_{65} p_{56}.
\end{equation}
Find \((q_3, q_4, q_5)\) by solving \eqref{eq:spherical_elbow}.

\subsubsection{2R-3R\textbar\textbar-2R Arm}
This robot is the limit of a 2R-3R-2R arm where the intersection point of the elbow joint moves to infinity.
We can write
\begin{equation}\label{eq:2R_shoulder_theta_S}
    R_{01} R_{12} h_3 = \rot(n_{SEW}, \theta_S) e_{SW}.
\end{equation}
Combining with the position equation gives
\begin{equation}
    h_3\tr (p_{23} + p_{34} + p_{45} + p_{56}) = h_3 \tr R_{02}\tr p_{07}.
\end{equation}
Solve for \(\theta_S\in [0, \pi]\) using Subproblem~4:
\begin{equation}
    e_{SW}\tr \rot(n_{SEW}, \theta_S)\tr p_{07} =   h_3\tr (p_{23} + p_{34} + p_{45} + p_{56}).
\end{equation}
Find \((q_1, q_2)\) using Subproblem~2:
\begin{equation}
    R_{12} h_3 = R_{10}\rot(n_{SEW}, \theta_S) e_{SW}.
\end{equation}
Find \((q_3+q_4+q_5, q_6, q_7)\) by solving a spherical joint:
\begin{equation} \label{eq:3_parallel_spherical_joint}
    R_{25} R_{56} R_{67} = R_{02}\tr R_{07}.
\end{equation}
Use Subproblem~3 to find \(q_4\):
\begin{equation}
    \norm{p_{34} + R_{34} p_{45}}
    = \norm{R_{02} \tr p_{07} - p_{23} - R_{25}p_{56}}.
\end{equation}
Use Subproblem~1 to find \(q_3\):
\begin{equation}
    R_{23}(p_{34} + R_{34} p_{45})
    = R_{02} \tr p_{07} - p_{23} - R_{25}p_{56}.
\end{equation}
Find \(q_5\) with subtraction, wrapping to \([-\pi, \pi]\) if desired.
\subsubsection{\texorpdfstring{R-R-3R\textbar\textbar\textsuperscript{E}-2R Arm}{R-R-3R\textbar\textbarᴱ-2R Arm}}
This is a more general version of a 2R-3R\textbar\textbar-2R robot. Use Subproblem 6 to find \((\theta_S, q_2)\), where \(\theta_S \in [0, \pi]\):
\begin{subequations}
\begin{multline}
        e_{SW} \tr \rot(n_{SEW}, \theta_S)\tr p_{07} - h_3 \tr R_{21} p_{12} \\
            = h_3\tr (p_{23} + p_{34} + p_{45} + p_{56}),
\end{multline}
\begin{equation}
        h_1 \tr \rot(n_{SEW}, \theta_S)\ e_{SW} - h_1 \tr R_{12} h_3= 0.
\end{equation}
\end{subequations}
Use Subproblem~1 to find \(q_1\) solving \eqref{eq:2R_shoulder_theta_S}. Find \((q_3+q_4+q_5, q_6, q_7)\) by solving a spherical joint \eqref{eq:3_parallel_spherical_joint}.
Use Subproblem~3 to find \(q_4\):
\begin{multline}
    \norm{p_{34} + R_{34} p_{45}}\\
    = \norm{R_{02} \tr p_{07} - R_{21} p_{12} - p_{23} - R_{25}p_{56}}.
\end{multline}
Use Subproblem~1 to find \(q_3\):
\begin{multline}
    R_{23}(p_{34} + R_{34} p_{45})\\
       = R_{02} \tr p_{07} - R_{21} p_{12} - p_{23} - R_{25}p_{56}.
\end{multline}
Find \(q_5\) with subtraction, wrapping to \([-\pi, \pi]\) if desired.

\subsubsection{\texorpdfstring{R-2R-2R\textsuperscript{E}-2R Arm (1D Search)}{R-2R-2Rᴱ-2R Arm (1D Search)}}
\begin{figure}[t]
    \centering
    \includegraphics[clip, width=\linewidth]{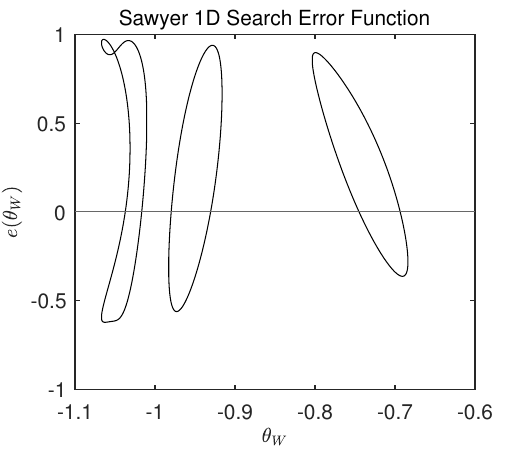}
    \caption{Error function for a Sawyer (R-2R-2R\textsuperscript{E}-2R) arm. This function has eight branches, although only four are seen for this pose. A 1D search may be used to find the zeros, which correspond to IK solutions.}
    \label{fig:sawyer_error_plot}
\end{figure}

\begin{figure}[t]
    \centering
    \subfloat[\(\theta_W = -1.0371\).]{\includegraphics[scale = 0.30, clip, trim = {2.05in 1in 0.85in 0.75in}]{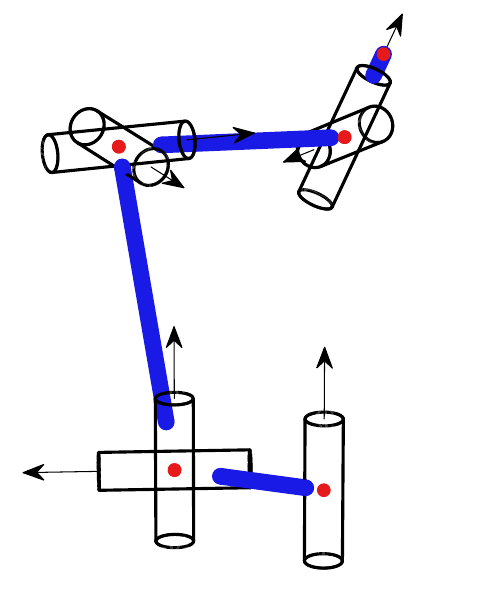}}%
    \subfloat[\(\theta_W = -1.0166\).]{\includegraphics[scale = 0.30, clip, trim = {2.05in 1in 0.85in 0.75in}]{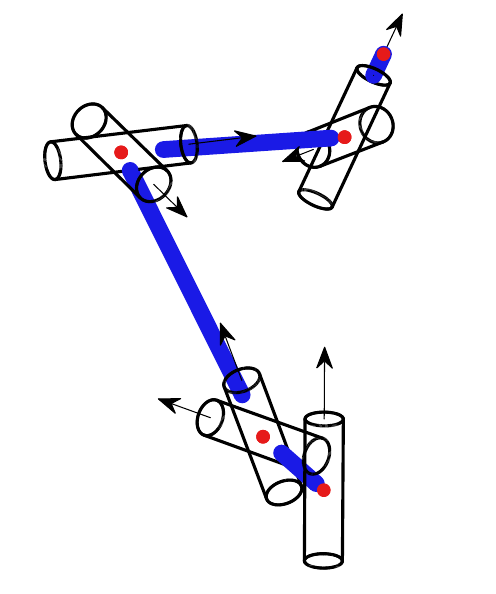}}%
    \subfloat[\(\theta_W = -0.9797\).]{\includegraphics[scale = 0.30, clip, trim = {2.05in 1in 0.85in 0.75in}]{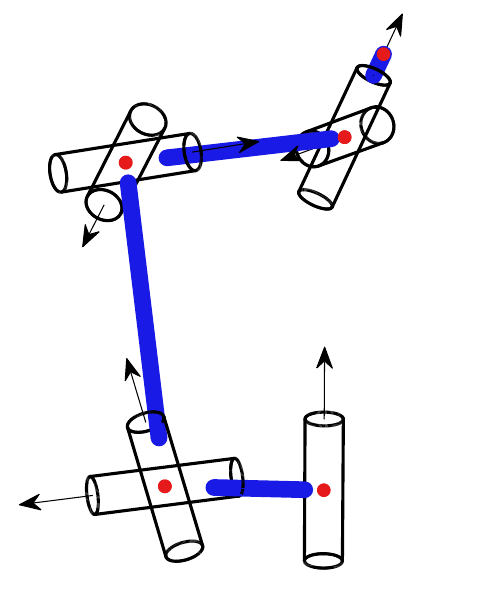}}\\
    \subfloat[\(\theta_W = -0.9304\).]{\includegraphics[scale = 0.30, clip, trim = {2.05in 1in 0.85in 0.75in}]{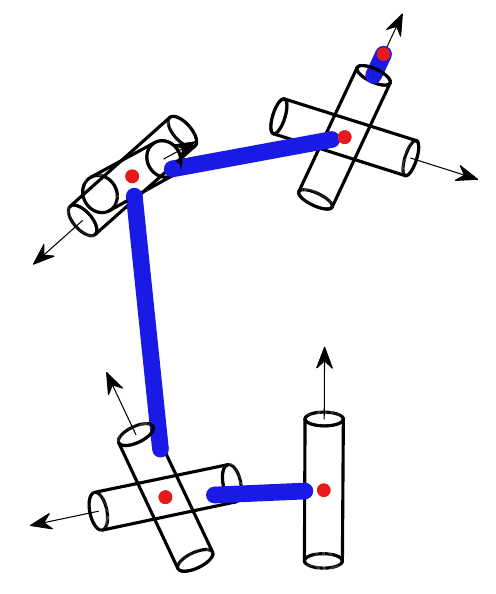}}%
    \subfloat[\(\theta_W = -0.7450\).]{\includegraphics[scale = 0.30, clip, trim = {2.05in 1in 0.85in 0.75in}]{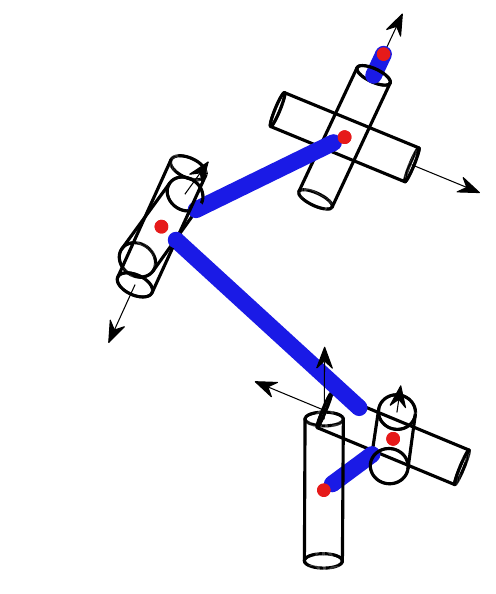}}%
    \subfloat[\(\theta_W = -0.6936\).]{\includegraphics[scale = 0.30, clip, trim = {2.05in 1in 0.85in 0.75in}]{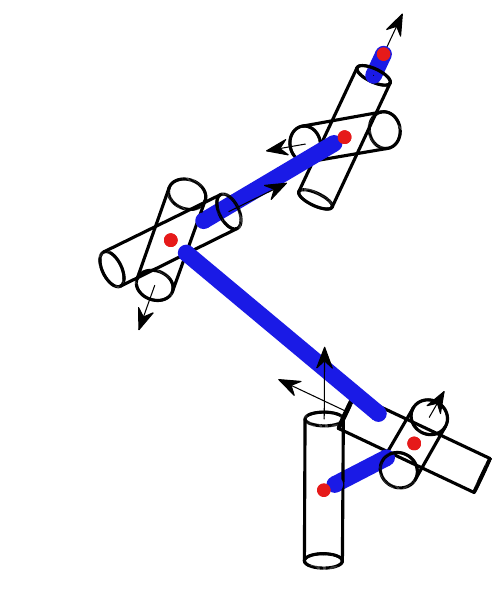}}%
    \caption{Six IK solutions for a Sawyer (R-2R-2R\textsuperscript{E}-2R) arm corresponding to the zeros of the error shown in Figure~\ref{fig:sawyer_error_plot}.}
    \label{fig:sawyer_IK}
\end{figure}

Given \(\theta_W\), we can write
\begin{equation}
    R_{05}p_{56} = \rot(n_{SEW}, \theta_W) e_{SW}\norm{p_{56}}.
\end{equation}
Then, find \(q_1\) using Subproblem~3:
\begin{equation} \label{eq:sawyer_q1_sp3}
    \norm{R_{01} p_{12} - p_{07} +  R_{05}p_{56}} = \norm{p_{34}}.
\end{equation}
Find \((q_2, q_3)\) using Subproblem~2:
\begin{equation} \label{eq:sawyer_q2_q3_sp2}
    R_{23}p_{34} = R_{21}(R_{10}p_{07}- R_{10}R_{05}p_{56} - p_{12}).
\end{equation}
Find \((q_4, q_5)\) using Subproblem~2:
\begin{equation}
    R_{45}p_{56} = R_{43}(R_{32}R_{21}(R_{10}p_{07}-p_{12})-p_{34}).
\end{equation}
The error is a metric of solvability of
\begin{equation}
    R_{56}R_{67} = R_{05}\tr R_{07}.
\end{equation}
By projecting onto \(h_6\) and \(h_7\), we get the error
\begin{equation}
    e(\theta_W) = h_6\tr R_{05}\tr R_{07}h_7 - h_6 \tr h_7.
\end{equation}
Search over \(\theta_W \in [-\pi,0]\) to find all solutions of \(e(\theta_W)=0\). For each solution \(\theta_W\), calculate \(q_6\) and \(q_7\) using Subproblem~1:
\begin{equation}
    R_{56} h_7 = R_{05}\tr R_{07}h_7,
\end{equation}
\begin{equation}
    R_{76} h_6 =  R_{07}\tr R_{05}h_6.
\end{equation}
A Sawyer arm example is shown in Figures~\ref{fig:sawyer_error_plot} and \ref{fig:sawyer_IK} \new{(Extension~2)}.

\subsubsection{\texorpdfstring{3R-R\textsuperscript{E}-2R-R Arm (1D Search)}{3R-Rᴱ-2R-R Arm (1D Search)}}
Given \(\theta_S \in [0,\pi]\), we have
\begin{equation}
    R_{03}p_{34} = \rot(n_{SEW},\theta_S)e_{SW}\norm{p_{34}}.
\end{equation}
Use Subproblem~3 to find \(q_7\) from
\begin{equation}
    \norm{R_{07}\tr(p_{07} - R_{03}p_{34}) - R_{76} p_{67}} = \norm{p_{45}}.
\end{equation}
Then, solve \((q_5, q_6)\) using Subproblem~2:
\begin{equation}
    R_{54} p_{45} = R_{56} (R_{67}R_{07}\tr(p_{07}-{R_{03}p_{34}})-p_{67}).
\end{equation}
We have the identity
\begin{equation}
    R_{43} p_{34} = R_{47} R_{07}\tr R_{03}p_{34},
\end{equation}
which may be expressed as an error function in terms of \(\theta_S\):
\begin{equation}
    e(\theta_S) = h_4\tr ( p_{34} -R_{47} R_{07}\tr R_{03}p_{34}).
\end{equation}
Search over \(\theta_S\in[0,\pi]\) to find the zeros of this error. For each solution \(\theta_S\), find \(q_4\) with Subproblem~1, then find \((q_1, q_2, q_3)\) by solving the spherical shoulder:
\begin{equation}
    R_{01}R_{12}R_{23} = R_{07}(R_{34}R_{45}R_{56}R_{67})\tr.
\end{equation}

\subsubsection{\texorpdfstring{R-2R\textsuperscript{S}-R-3R Arm (1D Search)}{R-2Rˢ-R-3R Arm (1D Search)}}
Given \(q_1 \in [-\pi, \pi]\), solve for \(q_4\) using Subproblem~3:
\begin{equation}
    \norm{p_{34} + R_{34}p_{45}} = \norm{R_{10} p_{07} - p_{12}}.
\end{equation}
Solve for \((q_2, q_3)\) using Subproblem~2:
\begin{equation}
    R_{21}(R_{10} p_{07} - p_{12}) = R_{23}(p_{34}+R_{34}p_{45}).
\end{equation}
Find the shoulder-elbow vector \(p_{SE} = R_{03}p_{34}\),
from which we can calculate the error for \(\psi\) in terms of \(q_1\).
A plot of this error is a good graphical tool to determine what range of SEW angle is feasible for a given end effector pose.
For each solution \(q_1\), find \((q_5,q_6,q_7)\) by solving \eqref{eq:spherical_wrist_with_RHS}.

\subsubsection{General 7-DOF Arm (2D Search)}
Pick \(\mathcal O_1 = \mathcal O_S\), \(\mathcal O_4 = \mathcal O_E\), and \(\mathcal O_7 = \mathcal O_W\).
\new{The shoulder, elbow, and wrist may be be placed arbitrarily as in Figure~\ref{fig:arbitrary_7_dof} and the solution will remain similar.}
Given \((q_1, q_2)\), find \(q_3\) using Subproblem~4, keeping only solutions which place the elbow in the correct half plane:
\begin{equation}
    n_{SEW}\tr R_{02} R_{23} p_{34} = -n_{SEW}\tr (R_{01}p_{12} + R_{02}p_{23}).
\end{equation}
Then, find \((q_5, q_6, q_7)\) with Subproblem 5 to solve
\begin{equation}
-p_{67} + R_{67} R_{07}\tr (p_{07}-p_{14}) = R_{65} (p_{56} + R_{54} p_{45}),
\end{equation}
and find the error
\begin{equation}
    e(q_1,q_2) = \norm{R_{03}\tr R_{07} R_{47}\tr h_4 - h_4}.
\end{equation}
Search over \((q_1, q_2)\) to find zeros of this error. Then, find \(q_4\) using Subproblem~1:
\begin{equation}
    R_{34}p = R_{03}\tr R_{07} R_{47}\tr p,
\end{equation}
where \(p\) is any vector not collinear with \(h_4\).
\subsection{\new{Continuous and }Least-Squares \new{IK}\cut{Inverse Kinematics}}
\cut{Any IK solution which only uses Subproblems~1--4 is robust to boundary singularities if the least-squares solutions for these subproblems is used}\new{Using the subproblem decomposition approach means that continuous non-exact solutions for \(q\) are returned even for branches where the solution does not exist}, as discussed in \cite*{elias2024canonical}. \new{Therefore, IK is numerically stable for paths where the robot switches branches by passing through a boundary singularity.}

For some special cases, the IK solution is also the solution to the global least-squares IK problem, which is posed as
\begin{equation} \label{eq:IK_LS_minimization}
\begin{aligned}
    \min_q & \norm{p_{0T}(q) - p_{0T}^{des}}\\
    \text{ s.t. } &R_{0T}(q) = R_{0T}^{des},\ \psi(q) = \psi^{des}
\end{aligned}
\end{equation}
where \((R_{0T}^{des}, p_{0T}^{des},\psi^{des})\) is the desired end effector pose and SEW angle. For some robots, the \( \psi(q) = \psi^{des}\) constraint cannot be achieved when at a boundary singularity because the self-motion manifold degenerates to a point, and so this constraint must be dropped when \(p_{0T}(q) \neq p_{0T}^{des}\).

3R-R-3R, 2R-3R-2R, and 2R-2R-3R arms can all achieve least-squares inverse kinematics if the task frame is at the wrist center (\(p_{7T} = 0\)), all 2R joints have a spherical workspace (\(h_i\tr h_{i+1}=h_{i+1}\tr p_{i+1,i+2}=0\)) and all 3R joints can achieve any orientation (\(h_i\tr h_{i+1}=h_{i+1}\tr h_{i+2}=0\)).

For a 3R-R-3R arm, since rotation of the whole arm is always possible between the spherical shoulder and wrist, the self-motion manifold does not degenerate to a point at boundary singularities. Therefore, \(\calO_E\) should be placed such that it is not collinear with \(\calO_S\) and \(\calO_W\) at the workspace boundary. (Using \(p_{SE} = e_{SE} = R_{03}h_4\) may also be a good option.)

For 2R-3R-2R and 2R-2R-3R arms, the robot's self-motion manifold degenerates into a point at boundary singularities, so although the SEW angle will be undefined at the boundary singularity, the SEW angle is not needed. This also means the SEW angle constraint in \eqref{eq:IK_LS_minimization} must be dropped.
\new{
\subsection{Polynomial Method}\label{sec:polynomial_method}
The tangent half-angle substitution \(x_i = \tan(q_i / 2)\) can be used to convert a search-based solutions into a system of multivariate polynomials. By eliminating all but one variable, we obtain a high-order polynomial in the tangent half-angle of one joint. After finding roots of this polynomial, the remaining joint angles are found in closed-form. Although this procedure is not as computationally efficient as 1D or 2D search, it does give a stronger guarantee of finding all solutions.
We demonstrate solving IK using the polynomial method for the Sawyer (R-2R-2R\textsuperscript{E}-2R) arm according to the procedure described in \cite*{elias2024canonical}.

To improve computational performance, we reduce the number of variables and search over a joint angle rather than \(\theta_W\). The downside is that we must check for extraneous solutions where the elbow is in the wrong half-plane. 

Given \(q_7\), find \(q_6\) using Subproblem~4:
\begin{equation}
    (R_{67} R_{07}\tr n_{SEW})\tr   R_{65} p_{56} = n_{SEW}\tr p_{07}.
\end{equation}
Then, find \(q_1\) with Subproblem~3 according to \eqref{eq:sawyer_q1_sp3},
where \(R_{05} = R_{07}  R_{76}  R_{65}\). Although Subproblem~2 could be used to find \(q_2\) and \(q_3\) using \eqref{eq:sawyer_q2_q3_sp2}, we instead first solve for \(q_2\) with Subproblem~4 to solve
\begin{equation}
    h_3\tr R_{21} (R_{10}p_{07} -R_{10}R_{05}p_{56} -p_{12}) = h_3\tr p_{34},
\end{equation}
and then find \(q_3\) with Subproblem~1. This allows us to eliminate \(x_3\) from the system of polynomials.
The error is
\begin{equation}
    e(q_7) = h_4\tr R_{03}\tr R_{07} R_{57}\tr h_5 - h_4\tr h_5.
\end{equation}
For all solutions of \(e(q_7) = 0\), we can find \(q_4\) and \(q_5\) with Subproblem~1:
\begin{align}
    R_{34} h_5 &= R_{03}\tr R_{07} R_{57}\tr h_5,\\
    R_{54} h_4 &= R_{57} R_{07}\tr R_{03} h_4. 
\end{align}
}

\begin{table*} \setlength{\tabcolsep}{4pt} 
    \small\sf\centering
    \caption{\new{Sawyer IK solutions using the polynomial method.}}
    \newoutline{\begin{tabular}{ c *{7}{c}}
        \toprule
        \# & \(q_1\) & \(q_2\) & \(q_3\) & \(q_4\) & \(q_5\) & \(q_6\) & \(q_7\) \\ \midrule
1& 0.7012115792 & -0.9732888736 & -0.09318675442 &   1.466219046 &   1.023549438 & -0.7523604269 & -0.8108011807\\
2& -1.187806104 &  -2.406581118 &    2.111970078 &   1.816987670 &   1.723460652 & -0.7764631130 & -0.7042361521\\
3&-0.4801904691 &  -1.230875621 &   -2.301720627 &  -2.019222054 &  -2.695866355 & -0.8165545740 & -0.5807494539\\
4& -2.104051752 &  -2.319400366 &  -0.7687046831 & -0.5435788511 &   2.572212359 &  0.7314410389 &  0.9764868428\\
5& 0.7028860908 &  -1.034458755 &  0.05293672172 &  0.9219195962 &  -1.476315039 &  0.7522268563 &   1.404840771\\
6& -1.439122724 &  -2.605604387 &    1.821941574 &  0.9918815495 & -0.4713994287 &  0.7552919261 &   1.423570856\\
7&-0.2361394798 &  -1.013327345 &   -2.064532180 &  -1.375427168 &   1.007651470 &  0.8154933152 &   1.682578759\\
        \bottomrule
    \end{tabular}}

    \label{tab:IK_solns}
\end{table*}
\setlength{\tabcolsep}{6pt} 

\new{
The Sawyer kinematics parameters (in mm) are:
\begin{equation}
\begin{gathered}
    h_1 = [0\ 0\ 1]\tr,\quad
    h_2 = h_4 = h_6 = [0\ 1\ 0]\tr,\\
    h_3 = h_5 = h_7 = [1\ 0\ 0]\tr,\\
    p_{12} = [81 \ 192.5 \ 0]\tr,\
    p_{23} = p_{45} = p_{67} = 0,\\
    p_{34} = [400 \ {-168.5} \ 0]\tr,\
    p_{56} = [400 \ 136.3 \ 0]\tr.
\end{gathered}
\end{equation}
Using the conventional SEW angle, we pick the following example pose (in mm):
\begin{equation}
\begin{gathered}
    R_{06}  = I_3, \quad p_{06}  = [500 \ 500 \  250]\tr,\\
    \psi  = 0, \quad  e_r  =  [0\ 0\ 1]\tr.
\end{gathered}
\end{equation}
The stereographic SEW angle can be used just as easily since the solution only depends on \(n_{SEW}\).

After converting all subproblems and the error equation using the tangent half-angle identity, we find four equations in four unknowns with \((23, 42, 5, 291)\) terms, respectively:
\begin{equation}
\begin{alignedat}{2}
P_1(x_1, x_6, x_7) &= 0,\quad  P_2(x_1, x_2, x_6, x_7) &&= 0,\\
P_3(x_6, x_7) &= 0,\quad  P_4(x_1, x_2, x_6, x_7) &&= 0.
\end{alignedat}
\end{equation}
Eliminating all variables but \(x_7\), we find a resultant univariate polynomial of degree 48. Of the 16 real solutions, 7 correspond to solutions to the IK problem, and the other 9 solutions correspond to poses with \(\psi = \pi\). All 7 IK solutions for this pose are shown in Table~\ref{tab:IK_solns} to 10 significant figures.
}

\section{Conclusion}\label{sec:conclusion}
We have introduced the general SEW angle which allows us to analyze the behavior of the conventional SEW angle but with an arbitrary reference direction function. A special choice of the reference direction function, the stereographic SEW angle, reduces the effect of the coordinate singularity as compared to the conventional SEW angle. The stereographic SEW angle allows the use of more of the workspace without encountering singularities. Even at a singularity, the arm can have continuous joint movement as long as a smooth path is taken directly through the singularity. We have shown that since an algorithmic singularity is unavoidable for any choice of parameterization of the redundant degree of freedom, the stereographic SEW angle is ideal in that it only encounters a singularity when the wrist is at a half-line from the shoulder.

We have also used the subproblem decomposition method to provide IK solutions for most known 7R robots. These solutions are often closed-form and may sometimes require a 1D or 2D search. This method finds all IK solutions and finds least-squares solutions for some robots as well. We provide IK solutions for both common robots as well as robots which do not seem to be manufactured yet, such as an R-R-3R\textbar\textbar-2R arm. \new{Furthermore, we are the first to demonstrate solving IK by finding a high-order polynomial in the tangent half-angle for 7R arms parameterized by the general SEW angle.}

\cut{There exist other 7R robots which may have similar solutions to the ones provided above. It may be worthwhile to find a closed-form or polynomial root-finding IK solution to the robots which are solved in this paper using a 1D or 2D search.}

\new{In the future, it would be interesting to further investigate resolving redundancy given this new parameterization, that is, picking the ideal trajectory for the stereographic SEW angle. It may also be worthwhile to further investigate IK for more types of 7R robots using the subproblem decomposition approach.}


\begin{dci} 
The authors declare that there is no conflict of interest.
\end{dci}
\begin{funding} 
The author(s) received no financial support for the research, authorship, and/or publication of this article.
\end{funding}

\interlinepenalty=10000 
\bibliographystyle{SageH}
\bibliography{bib, robot_kin_refs}

\begin{thebibliography}{81}
\providecommand{\natexlab}[1]{#1}
\providecommand{\url}[1]{\texttt{#1}}
\providecommand{\urlprefix}{URL }
\expandafter\ifx\csname urlstyle\endcsname\relax
  \providecommand{\doi}[1]{DOI:\discretionary{}{}{}#1}\else
  \providecommand{\doi}{DOI:\discretionary{}{}{}\begingroup \urlstyle{rm}\Url}\fi

\bibitem[{{ABB}(2022)}]{ABB_YUMI}
{ABB} (2022) {IRB 14000 YuMi} - collaborative robot.
\newblock \url{https://new.abb.com/products/robotics/collaborative-robots/yumi/irb-14000-yumi}.
\newblock (accessed Oct. 31, 2022).

\bibitem[{An et~al.(2014)An, Clement and Reed}]{an2014analytical}
An HH, Clement WI and Reed B (2014) Analytical inverse kinematic solution with self-motion constraint for the 7-{DOF} restore robot arm.
\newblock In: \emph{2014 IEEE/ASME International Conference on Advanced Intelligent Mechatronics}. IEEE, pp. 1325--1330.

\bibitem[{Baillieul(1985)}]{baillieul1985kinematic}
Baillieul J (1985) Kinematic programming alternatives for redundant manipulators.
\newblock In: \emph{Proceedings. 1985 IEEE International Conference on Robotics and Automation}, volume~2. IEEE, pp. 722--728.

\bibitem[{Baillieul et~al.(1987)Baillieul, Hollerbach, Brockett, Martin, Percy and Thomas}]{baillieul1987kinematically}
Baillieul J, Hollerbach J, Brockett R, Martin D, Percy R and Thomas R (1987) Kinematically redundant robot manipulators.
\newblock In: \emph{Proceedings of the Workshop on Space Telerobotics}, volume~2. Jet Propulsion Lab., California Inst. of Tech., pp. 245--255.

\bibitem[{Boumans and Heemskerk(1998)}]{boumans1998european}
Boumans R and Heemskerk C (1998) The {European} robotic arm for the international space station.
\newblock \emph{Robotics and Autonomous systems} 23(1-2): 17--27.

\bibitem[{Burdick and Seraji(1989)}]{burdick1989characterization}
Burdick J and Seraji H (1989) Characterization and control of self-motions in redundant manipulators.
\newblock In: \emph{Proceedings of the NASA Conference on Space Telerobotics, Volume 2}. pp. 3--14.

\bibitem[{Carignan and Howard(2000)}]{carignan2000partitioned}
Carignan CR and Howard RD (2000) A partitioned redundancy management scheme for an eight-joint revolute manipulator.
\newblock \emph{Journal of Robotic Systems} 17(9): 453--468.

\bibitem[{Carignan et~al.(2001)Carignan, Lane and Churchill}]{carignan2001controlling}
Carignan CR, Lane JC and Churchill PJ (2001) Controlling robots on-orbit.
\newblock In: \emph{Proceedings 2001 IEEE International Symposium on Computational Intelligence in Robotics and Automation (Cat. No. 01EX515)}. IEEE, pp. 314--319.

\bibitem[{Chen et~al.(2023)Chen, Zhang, Huang, Wu and Ota}]{chen2023kinematics}
Chen Y, Zhang X, Huang Y, Wu Y and Ota J (2023) Kinematics optimization of a novel 7-{DOF} redundant manipulator.
\newblock \emph{Robotics and Autonomous Systems} 163: 104377.

\bibitem[{Chiaverini(1997)}]{chiaverini1997singularity}
Chiaverini S (1997) Singularity-robust task-priority redundancy resolution for real-time kinematic control of robot manipulators.
\newblock \emph{IEEE Transactions on Robotics and Automation} 13(3): 398--410.

\bibitem[{Choi et~al.(2004)Choi, Oh, Oh, Park and Chung}]{choi2004multiple}
Choi Y, Oh Y, Oh SR, Park J and Chung WK (2004) Multiple tasks manipulation for a robotic manipulator.
\newblock \emph{Advanced Robotics} 18(6): 637--653.

\bibitem[{Crane~III et~al.(1991)Crane~III, Duffy and Carnahan}]{crane1991kinematic}
Crane~III CD, Duffy J and Carnahan T (1991) A kinematic analysis of the space station remote manipulator system ({SSRMS}).
\newblock \emph{Journal of Robotic Systems} 8(5): 637--658.

\bibitem[{Debus and Dougherty(2009)}]{debus2009overview}
Debus T and Dougherty S (2009) Overview and performance of the front-end robotics enabling near-term demonstration ({FREND}) robotic arm.
\newblock In: \emph{AIAA Infotech@Aerospace Conference and AIAA Unmanned...Unlimited Conference}. p. 1870.

\bibitem[{Dupuis(2001)}]{dupuis2001general}
Dupuis E (2001) \emph{A general framework for the manual teleoperation of kinematically redundant space-based manipulators}.
\newblock PhD Thesis, McGill University.

\bibitem[{Elias and Wen(2024)}]{elias2024canonical}
Elias AJ and Wen JT (2024) {IK-Geo}: Unified robot inverse kinematics using subproblem decomposition.
\newblock \emph{arXiv preprint arXiv:2211.05737v3} .

\bibitem[{{FANUC}(2022)}]{fanuc}
{FANUC} (2022) {FANUC R-1000iA/120F-7B} robot.
\newblock \url{https://www.fanucamerica.com/products/robots/series/r-1000ia/r-1000ia-120f-7b-7-axis-robot}.
\newblock (accessed Dec. 21, 2022).

\bibitem[{Faria et~al.(2018)Faria, Ferreira, Erlhagen, Monteiro and Bicho}]{faria2018position}
Faria C, Ferreira F, Erlhagen W, Monteiro S and Bicho E (2018) Position-based kinematics for 7-{DoF} serial manipulators with global configuration control, joint limit and singularity avoidance.
\newblock \emph{Mechanism and Machine Theory} 121: 317--334.

\bibitem[{Flacco et~al.(2012)Flacco, De~Luca and Khatib}]{flacco2012motion}
Flacco F, De~Luca A and Khatib O (2012) Motion control of redundant robots under joint constraints: Saturation in the null space.
\newblock In: \emph{2012 IEEE International Conference on Robotics and Automation}. IEEE, pp. 285--292.

\bibitem[{{Franka Emika}(2022)}]{franka}
{Franka Emika} (2022) Franka production 3.
\newblock \url{https://www.franka.de/production}.
\newblock (accessed Dec. 21, 2022).

\bibitem[{Gong et~al.(2019)Gong, Li and Zhang}]{gong2019analytical}
Gong M, Li X and Zhang L (2019) Analytical inverse kinematics and self-motion application for 7-{DOF} redundant manipulator.
\newblock \emph{IEEE Access} 7: 18662--18674.

\bibitem[{Gottlieb(1986)}]{gottlieb1986robots}
Gottlieb DH (1986) Robots and fibre bundles.
\newblock \emph{Bull. Soc. Math. Belg} 38: 219--223.

\bibitem[{Hollerbach(1985)}]{hollerbach1985optimum}
Hollerbach JM (1985) Optimum kinematic design for a seven degree of freedom manipulator.
\newblock In: \emph{Robotics research: The second international symposium}. Citeseer, pp. 215--222.

\bibitem[{Hollerbach and Suh(1987)}]{hollerbach1987redundancy}
Hollerbach JM and Suh K (1987) Redundancy resolution of manipulators through torque optimization.
\newblock \emph{IEEE Journal on Robotics and Automation} 3(4): 308--316.

\bibitem[{Hopf(1927)}]{hopf1927vektorfelder}
Hopf H (1927) Vektorfelder inn-dimensionalen mannigfaltigkeiten.
\newblock \emph{Mathematische Annalen} 96(1): 225--249.

\bibitem[{Jiang et~al.(2013)Jiang, Huo, Liu and Liu}]{jiang2013integrated}
Jiang L, Huo X, Liu Y and Liu H (2013) An integrated inverse kinematic approach for the 7-{DOF} humanoid arm with offset wrist.
\newblock In: \emph{2013 IEEE International Conference on Robotics and Biomimetics (ROBIO)}. IEEE, pp. 2737--2742.

\bibitem[{Jin et~al.(2020)Jin, Liu, Wang and Liu}]{jin2020efficient}
Jin M, Liu Q, Wang B and Liu H (2020) An efficient and accurate inverse kinematics for 7-{DOF} redundant manipulators based on a hybrid of analytical and numerical method.
\newblock \emph{IEEE Access} 8: 16316--16330.

\bibitem[{Kim et~al.(2011)Kim, Miller, Al-Refai, Brand and Rosen}]{kim2011redundancy}
Kim H, Miller LM, Al-Refai A, Brand M and Rosen J (2011) Redundancy resolution of a human arm for controlling a seven {DOF} wearable robotic system.
\newblock In: \emph{2011 Annual International Conference of the IEEE Engineering in Medicine and Biology Society}. IEEE, pp. 3471--3474.

\bibitem[{Kim et~al.(2012)Kim, Miller, Byl, Abrams and Rosen}]{kim2012redundancy}
Kim H, Miller LM, Byl N, Abrams GM and Rosen J (2012) Redundancy resolution of the human arm and an upper limb exoskeleton.
\newblock \emph{IEEE Transactions on Biomedical Engineering} 59(6): 1770--1779.

\bibitem[{Kreutz-Delgado et~al.(1990)Kreutz-Delgado, Long and Seraji}]{kreutz1990kinematic}
Kreutz-Delgado K, Long M and Seraji H (1990) Kinematic analysis of 7 {DOF} anthropomorphic arms.
\newblock In: \emph{Proceedings., IEEE International Conference on Robotics and Automation}. IEEE, pp. 824--830.

\bibitem[{Kreutz-Delgado et~al.(1992)Kreutz-Delgado, Long and Seraji}]{kreutz1992kinematic}
Kreutz-Delgado K, Long M and Seraji H (1992) Kinematic analysis of 7-{DOF} manipulators.
\newblock \emph{The International Journal of Robotics Research} 11(5): 469--481.

\bibitem[{{KUKA}(2022)}]{KUKA}
{KUKA} (2022) {LBR} iiwa.
\newblock \url{https://www.kuka.com/en-us/products/robotics-systems/industrial-robots/lbr-iiwa}.
\newblock (accessed Oct. 31, 2022).

\bibitem[{Lamperti et~al.(2015)Lamperti, Zanchettin and Rocco}]{lamperti2015redundancy}
Lamperti C, Zanchettin AM and Rocco P (2015) A redundancy resolution method for an anthropomorphic dual-arm manipulator based on a musculoskeletal criterion.
\newblock In: \emph{2015 IEEE/RSJ International Conference on Intelligent Robots and Systems (IROS)}. IEEE, pp. 1846--1851.

\bibitem[{Lee and Buss(2006)}]{lee2006redundancy}
Lee KK and Buss M (2006) Redundancy resolution with multiple criteria.
\newblock In: \emph{2006 IEEE/RSJ International Conference on Intelligent Robots and Systems}. IEEE, pp. 598--603.

\bibitem[{Li et~al.(2022)Li, Han, He, Li, Liu and Xiong}]{li2022human}
Li S, Han K, He P, Li Z, Liu Y and Xiong Y (2022) Human-like redundancy resolution: An integrated inverse kinematics scheme for anthropomorphic manipulators with radial elbow offset.
\newblock \emph{Advanced Engineering Informatics} 54: 101812.

\bibitem[{Li et~al.(2013)Li, Roldan, Milutinovi{\'c} and Rosen}]{li2013rotational}
Li Z, Roldan JR, Milutinovi{\'c} D and Rosen J (2013) The rotational axis approach for resolving the kinematic redundancy of the human arm in reaching movements.
\newblock In: \emph{2013 35th Annual International Conference of the IEEE Engineering in Medicine and Biology Society (EMBC)}. IEEE, pp. 2507--2510.

\bibitem[{Liu et~al.(2017)Liu, Chen and Steil}]{liu2017analytical}
Liu W, Chen D and Steil J (2017) Analytical inverse kinematics solver for anthropomorphic 7-{DOF} redundant manipulators with human-like configuration constraints.
\newblock \emph{Journal of Intelligent \& Robotic Systems} 86(1): 63--79.

\bibitem[{Ma et~al.(2021)Ma, Xie, Jiang and Liu}]{ma2021precise}
Ma B, Xie Z, Jiang Z and Liu H (2021) Precise semi-analytical inverse kinematic solution for 7-{DOF} offset manipulator with arm angle optimization.
\newblock \emph{Frontiers of Mechanical Engineering} 16(3): 435--450.

\bibitem[{Marani et~al.(2003)Marani, Kim, Yuh and Chung}]{marani2003algorithmic}
Marani G, Kim J, Yuh J and Chung WK (2003) Algorithmic singularities avoidance in task-priority based controller for redundant manipulators.
\newblock In: \emph{Proceedings 2003 IEEE/RSJ International Conference on Intelligent Robots and Systems (IROS 2003)(Cat. No. 03CH37453)}, volume~4. IEEE, pp. 3570--3574.

\bibitem[{McGrath(2016)}]{mcgrath2016extremely}
McGrath P (2016) An extremely short proof of the hairy ball theorem.
\newblock \emph{The American Mathematical Monthly} 123(5): 502--503.

\bibitem[{{Motiv Space Systems}(2022)}]{osam2}
{Motiv Space Systems} (2022) {xLink} space-rated modular robotic arm system.
\newblock \url{https://motivss.com/products-capabilities/robotics/xlink/}.
\newblock (accessed Dec. 21, 2022).

\bibitem[{Mukherji et~al.(2001)Mukherji, Ray, Stieber and Lymer}]{mukherji2001special}
Mukherji R, Ray DA, Stieber M and Lymer J (2001) Special purpose dexterous manipulator ({SPDM}) advanced control features and development test results.
\newblock \emph{Proceedings of the 6th International Symposium on Artificial Intelligence, Robotics and Automation in Space (i-SAIRAS)} .

\bibitem[{Nammoto and Kosuge(2012)}]{nammoto2012analytical}
Nammoto T and Kosuge K (2012) An analytical solution for a redundant manipulator with seven degrees of freedom.
\newblock \emph{International Journal of Automation and Smart Technology} 2(4): 339--346.

\bibitem[{Naylor et~al.(2007)Naylor, Atkins and Roderick}]{naylor2007visual}
Naylor M, Atkins E and Roderick S (2007) Visual target recognition and tracking for autonomous manipulation tasks.
\newblock In: \emph{AIAA Guidance, Navigation and Control Conference and Exhibit}. p. 6323.

\bibitem[{Needham(1997)}]{needham1997visual}
Needham T (1997) \emph{Visual complex analysis}.
\newblock Oxford University Press.

\bibitem[{O'Neil and Chen(2000)}]{o2000instability}
O'Neil K and Chen YC (2000) Instability of pseudoinverse acceleration control of redundant mechanisms.
\newblock In: \emph{Proceedings 2000 ICRA. Millennium Conference. IEEE International Conference on Robotics and Automation. Symposia Proceedings (Cat. No. 00CH37065)}, volume~3. IEEE, pp. 2575--2582.

\bibitem[{Pfurner(2016)}]{pfurner2016closed}
Pfurner M (2016) Closed form inverse kinematics solution for a redundant anthropomorphic robot arm.
\newblock \emph{Computer Aided Geometric Design} 47: 163--171.

\bibitem[{Pieper(1969)}]{pieper1969kinematics}
Pieper DL (1969) \emph{The kinematics of manipulators under computer control}.
\newblock Stanford University.

\bibitem[{Poincar{\'e}(1885)}]{poincare1885courbes}
Poincar{\'e} H (1885) Sur les courbes d{\'e}finies par les {\'e}quations diff{\'e}rentielles.
\newblock \emph{J. Math. Pures Appl.} 4: 167--244.

\bibitem[{{Productive Robots}(2023)}]{OB7}
{Productive Robots} (2023) {OB7} collaborative robots.
\newblock \url{https://www.productiverobotics.com/ob7-products/}.
\newblock (accessed Jun. 21, 2023).

\bibitem[{Pryor(2023)}]{pryor2023teleoperation}
Pryor JWR (2023) \emph{Teleoperation Methods for High-Risk, High-Latency Environments}.
\newblock PhD Thesis, Johns Hopkins University.

\bibitem[{{Rethink Robotics}(2015)}]{baxter}
{Rethink Robotics} (2015) Baxter overview.
\newblock \url{https://sdk.rethinkrobotics.com/wiki/Baxter_Overview}.
\newblock (accessed Dec. 21, 2022).

\bibitem[{{Rethink Robotics}(2022)}]{sawyer}
{Rethink Robotics} (2022) Sawyer, the high performance collaborative robot.
\newblock \url{https://www.rethinkrobotics.com/sawyer}.
\newblock (accessed Dec. 21, 2022).

\bibitem[{{Robotics Research Corporation}(2005)}]{RRC}
{Robotics Research Corporation} (2005) Dexterous manipulators and advanced control systems.
\newblock Technical report, {Robotics Research Corporation}.
\newblock \urlprefix\url{http://www.robotics-research.com/RRCTechDoc.PDF}.
\newblock {Accessed Oct. 31, 2022}.

\bibitem[{Scott and Carignan(2008)}]{scott2008line}
Scott NA and Carignan CR (2008) A line-based obstacle avoidance technique for dexterous manipulator operations.
\newblock In: \emph{2008 IEEE International Conference on Robotics and Automation}. IEEE, pp. 3353--3358.

\bibitem[{Seraji(1989)}]{seraji1989configuration}
Seraji H (1989) Configuration control of redundant manipulators: Theory and implementation.
\newblock \emph{IEEE Transactions on Robotics and Automation} 5(4): 472--490.

\bibitem[{Shi et~al.(2021)Shi, Guo, Chen, Chen and Yang}]{shi2021kinematics}
Shi X, Guo Y, Chen X, Chen Z and Yang Z (2021) Kinematics and singularity analysis of a 7-{DOF} redundant manipulator.
\newblock \emph{Sensors} 21(21): 7257.

\bibitem[{Shimizu et~al.(2007)Shimizu, Yoon and Kitagaki}]{shimizu2007practical}
Shimizu M, Yoon WK and Kitagaki K (2007) A practical redundancy resolution for 7 {DOF} redundant manipulators with joint limits.
\newblock In: \emph{Proceedings 2007 IEEE International Conference on Robotics and Automation}. IEEE, pp. 4510--4516.

\bibitem[{Sinha and Chakraborty(2019)}]{sinha2019geometric}
Sinha A and Chakraborty N (2019) Geometric search-based inverse kinematics of 7-{DoF} redundant manipulator with multiple joint offsets.
\newblock In: \emph{2019 International Conference on Robotics and Automation (ICRA)}. IEEE, pp. 5592--5598.

\bibitem[{Stanczyk and Buss(2004)}]{stanczyk2004development}
Stanczyk B and Buss M (2004) Development of a telerobotic system for exploration of hazardous environments.
\newblock In: \emph{2004 IEEE/RSJ International Conference on Intelligent Robots and Systems (IROS)(IEEE Cat. No. 04CH37566)}, volume~3. IEEE, pp. 2532--2537.

\bibitem[{Stanczyk et~al.(2006)Stanczyk, Peer and Buss}]{stanczyk2006development}
Stanczyk B, Peer A and Buss M (2006) Development of a high-performance haptic telemanipulation system with dissimilar kinematics.
\newblock \emph{Advanced robotics} 20(11): 1303--1320.

\bibitem[{Su et~al.(2018{\natexlab{a}})Su, Enayati, Vantadori, Spinoglio, Ferrigno and De~Momi}]{su2018online}
Su H, Enayati N, Vantadori L, Spinoglio A, Ferrigno G and De~Momi E (2018{\natexlab{a}}) Online human-like redundancy optimization for tele-operated anthropomorphic manipulators.
\newblock \emph{International Journal of Advanced Robotic Systems} 15(6): 1729881418814695.

\bibitem[{Su et~al.(2018{\natexlab{b}})Su, Sandoval, Makhdoomi, Ferrigno and De~Momi}]{su2018safety_b}
Su H, Sandoval J, Makhdoomi M, Ferrigno G and De~Momi E (2018{\natexlab{b}}) Safety-enhanced human-robot interaction control of redundant robot for teleoperated minimally invasive surgery.
\newblock In: \emph{2018 IEEE International Conference on Robotics and Automation (ICRA)}. IEEE, pp. 6611--6616.

\bibitem[{Su et~al.(2018{\natexlab{c}})Su, Sandoval, Vieyres, Poisson, Ferrigno and De~Momi}]{su2018safety_a}
Su H, Sandoval J, Vieyres P, Poisson G, Ferrigno G and De~Momi E (2018{\natexlab{c}}) Safety-enhanced collaborative framework for tele-operated minimally invasive surgery using a 7-dof torque-controlled robot.
\newblock \emph{International Journal of Control, Automation and Systems} 16: 2915--2923.

\bibitem[{Su et~al.(2019)Su, Yang, Ferrigno and De~Momi}]{su2019improved}
Su H, Yang C, Ferrigno G and De~Momi E (2019) Improved human--robot collaborative control of redundant robot for teleoperated minimally invasive surgery.
\newblock \emph{IEEE Robotics and Automation Letters} 4(2): 1447--1453.

\bibitem[{Swaim et~al.(1994)Swaim, Arend, Bevill, Decker, Dunn, Read, Reiher, Richard, Ruta and Teplitz}]{swaim1994use}
Swaim PL, Arend JJ, Bevill PJ, Decker RJ, Dunn JC, Read DA, Reiher RE, Richard BJ, Ruta KJ and Teplitz S (1994) Use of manipulators in assembly of space station freedom.
\newblock In: Skaar SB (ed.) \emph{Teleoperation and robotics in space}, \emph{Progress in Astronautics and Aeronautics}, volume 161, chapter~16. Washington, DC: American Institute of Aeronautics and Astronautics, Inc., pp. 443--473.

\bibitem[{Tian et~al.(2021)Tian, Xu and Zhan}]{tian2021analytical}
Tian X, Xu Q and Zhan Q (2021) An analytical inverse kinematics solution with joint limits avoidance of 7-{DOF} anthropomorphic manipulators without offset.
\newblock \emph{Journal of the Franklin Institute} 358(2): 1252--1272.

\bibitem[{Tondu(2006)}]{tondu2006closed}
Tondu B (2006) A closed-form inverse kinematic modelling of a {7R} anthropomorphic upper limb based on a joint parametrization.
\newblock In: \emph{2006 6th IEEE-RAS International Conference on Humanoid Robots}. IEEE, pp. 390--397.

\bibitem[{Tsumaki et~al.(2001)Tsumaki, Fiorini, Chalfant and Seraji}]{tsumaki2001numerical}
Tsumaki Y, Fiorini P, Chalfant G and Seraji H (2001) A numerical {SC} approach for a teleoperated 7-{DOF} manipulator.
\newblock In: \emph{Proceedings 2001 ICRA. IEEE International Conference on Robotics and Automation (Cat. No. 01CH37164)}, volume~1. IEEE, pp. 1039--1044.

\bibitem[{Ufactory(2022)}]{xarm}
Ufactory (2022) Ufactory {xArm} 7.
\newblock \url{https://www.ufactory.cc/product-page/ufactory-xarm-7}.
\newblock (accessed Dec. 21, 2022).

\bibitem[{{Universal Robots}(2022)}]{UR5}
{Universal Robots} (2022) {UR5} collaborative robot arm.
\newblock \url{https://www.universal-robots.com/products/ur5-robot/}.
\newblock (accessed Oct. 31, 2022).

\bibitem[{Wang et~al.(2019)Wang, Peng, Hou, Li, Luo, Chen and Wang}]{wang2019kinematic}
Wang C, Peng L, Hou ZG, Li J, Luo L, Chen S and Wang W (2019) Kinematic redundancy analysis during goal-directed motion for trajectory planning of an upper-limb exoskeleton robot.
\newblock In: \emph{2019 41st Annual International Conference of the IEEE Engineering in Medicine and Biology Society (EMBC)}. IEEE, pp. 5251--5255.

\bibitem[{Wang and Artemiadis(2013)}]{wang2013closed}
Wang Y and Artemiadis P (2013) Closed-form inverse kinematic solution for anthropomorphic motion in redundant robot arms.
\newblock \emph{Advances in Robotics and Automation} 2(3).

\bibitem[{Wang et~al.(2021)Wang, Zhao, Wang, Zhang, Li and Liu}]{wang2021inverse}
Wang Y, Zhao C, Wang X, Zhang P, Li P and Liu H (2021) Inverse kinematics of a 7-{DOF} spraying robot with {4R} 3-{DOF} non-spherical wrist.
\newblock \emph{Journal of Intelligent \& Robotic Systems} 101(4): 1--17.

\bibitem[{Wang and Kazerounian(1995)}]{wang1995identification}
Wang Z and Kazerounian K (1995) Identification and resolution of structural and algorithmic singularity in redundancy control of serial manipulations.
\newblock \emph{Journal of robotic systems} 12(7): 465--478.

\bibitem[{Xiong et~al.(2020)Xiong, Zhou and Yao}]{xiong2020null}
Xiong G, Zhou Y and Yao J (2020) Null-space impedance control of 7-degree-of-freedom redundant manipulators based on the arm angles.
\newblock \emph{International Journal of Advanced Robotic Systems} 17(3): 1729881420925297.

\bibitem[{Xu et~al.(2014)Xu, She and Xu}]{xu2014analytical}
Xu W, She Y and Xu Y (2014) Analytical and semi-analytical inverse kinematics of {SSRMS}-type manipulators with single joint locked failure.
\newblock \emph{Acta Astronautica} 105(1): 201--217.

\bibitem[{Yan et~al.(2014)Yan, Mu and Xu}]{yan2014analytical}
Yan L, Mu Z and Xu W (2014) Analytical inverse kinematics of a class of redundant manipulator based on dual arm-angle parameterization.
\newblock In: \emph{2014 IEEE International Conference on Systems, Man, and Cybernetics (SMC)}. IEEE, pp. 3744--3749.

\bibitem[{Yaskawa(2022{\natexlab{a}})}]{moto50d}
Yaskawa (2022{\natexlab{a}}) Motoman {SIA50D} 7-axis robot arm.
\newblock \url{https://www.motoman.com/en-us/products/robots/industrial/assembly-handling/sia-series/sia50d}.
\newblock (accessed Dec. 21, 2022).

\bibitem[{Yaskawa(2022{\natexlab{b}})}]{moto5d}
Yaskawa (2022{\natexlab{b}}) Motoman {SIA5D} 7-axis robot arm.
\newblock \url{https://www.motoman.com/en-us/products/robots/industrial/assembly-handling/sia-series/sia5d}.
\newblock (accessed Dec. 21, 2022).

\bibitem[{Zhao et~al.(2023)Zhao, Yang, Zhao, Yang and Zhao}]{zhao2023inverse}
Zhao J, Yang X, Zhao Z, Yang G and Zhao L (2023) Inverse kinematics and multi-objective configuration optimization of the {SSRMS} manipulator.
\newblock \emph{Advances in Space Research} .

\bibitem[{Zhu et~al.(2021)Zhu, Wang and Ma}]{zhu2021design}
Zhu X, Wang X and Ma Y (2021) Design and development of teleoperation interactive system for 7-dof space redundant manipulator.
\newblock In: \emph{2021 5th International Conference on Automation, Control and Robots (ICACR)}. IEEE, pp. 179--183.

\end{thebibliography}


\new{\section*{Appendix. Index to multimedia extensions}
}

Multimedia extensions are available at \url{https://www.youtube.com/@AlexEliasRobotics}.

\begin{table}[H]
    \centering
    \small\sf \caption*{\new{Table of Multimedia Extensions}}
\newoutline{
\begin{tabularx}{0.953\columnwidth}{l l X}
    \toprule
     Extension & Media Type & Description\\ \midrule
     1 & Video &\raggedright Comparing conventional and stereographic SEW angles\arraybackslash\\[12pt]
     2 & Video &\raggedright  Sawyer inverse kinematics solutions using 1D search\arraybackslash\\\bottomrule
\end{tabularx}}
\end{table}

\end{document}